\documentclass{article}

\PassOptionsToPackage{numbers, compress}{natbib}

\usepackage[preprint]{neurips_2026}

\usepackage[utf8]{inputenc}
\usepackage[T1]{fontenc}
\usepackage{hyperref}
\usepackage{url}
\usepackage{booktabs}
\usepackage{amsfonts}
\usepackage{nicefrac}
\usepackage{microtype}
\usepackage{xcolor}

\usepackage{amsmath}
\usepackage{graphicx}
\usepackage{multirow}
\usepackage{multicol}
\usepackage{subcaption}
\usepackage[inline]{enumitem}
\usepackage{pifont}
\newcommand{\cmark}{\ding{51}}
\newcommand{\xmark}{\ding{55}}

\definecolor{darkred}{RGB}{139,0,0}
\definecolor{darkgreen}{RGB}{0,100,0}

\title{Mind the Gap? A Distributional Comparison of Real and Synthetic Priors for Tabular Foundation Models}

\author{%
  Alex O. Davies \\
  School of Geographical Sciences\\
  University of Bristol, UK\\
  \texttt{alexander.davies@bristol.ac.uk} \\
  \And
  Telmo de Menezes e Silva Filho \\
  School of Engineering Mathematics and Technology\\
  University of Bristol, UK\\
  \And
  Nirav Ajmeri \\
  School of Computer Science\\
  University of Bristol, UK\\
}

\begin{document}

\maketitle


\begin{abstract}
Tabular foundation models are pre-trained on one of three classes of corpus: curated datasets drawn from benchmark repositories, tables harvested at scale from the web, or synthetic tables sampled from a parametric generative prior.
Despite the centrality of pre-training data to model performance, little is known about how these corpora relate to one another in distribution, and the impact this has on downstream performance.
In this work we take three canonical, archetypal datasets used to train tabular foundation models; the T4 dataset represents web-scraped corpora, the TabFM dataset curated tables from Kaggle, and the TabICL dataset as the only well-used synthetic prior with publicly available parameters.
We characterise each corpus using aggregate features over whole tables, columns and correlations, and compare them using discriminator AUCs and k-NN coverage metrics.
We find that the TabICL synthetic prior occupies a narrow region of the space of real tables, that this mismatch cannot be closed by optimising prior hyper-parameters across more than 86 thousand configurations, and that curated and web-scraped corpora are broadly interchangeable on a distributional level in feature space.
Surprisingly, the distributional gap between synthetic pre-training data and real tables has a clearly detectable effect on performance under neither feature-based proximity measures or TabICL's own internal representations, suggesting that coverage of the real-data distribution is not the primary driver of TabICL's generalisation.
\end{abstract}

\section{Introduction}
\label{sec:introduction}

Tabular data is among the most prevalent data modalities in applied machine learning, spanning domains from healthcare and finance to scientific research and industry \cite{akoglu_rtm_2008, jiangRepresentationLearningTabular2026, shwartz-zivTabularDataDeep2022, quinnLiteratureReviewExplainable2024}. Unlike images or text, tabular data lacks a canonical structure, with datasets varying widely in the number of columns, the mix of numeric and categorical features, and the statistical properties of individual columns. This heterogeneity has historically made it difficult to transfer knowledge across datasets, and classical approaches typically require training a new model from scratch for each task.

Tabular foundation models (TFMs) offer a promising alternative. By pre-training on large corpora of tabular data, TFMs can perform in-context learning (ICL) at inference time --- generalising to new tasks from a small number of examples without further gradient updates \cite{hollmannAccuratePredictionsSmall2025, quTabICLTabularFoundation2025, wenSupervisedGenerativeNovel2024, zhangMitraMixedSynthetic2025}. 
These models are comparatively expensive to train as a result; TabPFN required 8 GPUs applied over two weeks train-time \cite{hollmannAccuratePredictionsSmall2025}.
The reward for this cost has been competitive performance against task-specific baselines \cite{hollmannAccuratePredictionsSmall2025,zhangMitraMixedSynthetic2025}, but its success depends critically on the quality and diversity of the pre-training corpus.

Three broad strategies have emerged for constructing these corpora. The first collates tables from existing benchmark repositories, as in TabFM \cite{wenSupervisedGenerativeNovel2024}. The second harvests tables at scale from the web, exemplified by T4 and Tabula-8B \cite{gardnerLargeScaleTransfer2024a}. The third forgoes real data entirely, instead sampling synthetic tables from a parametric generative prior; TabICL \cite{quTabICLTabularFoundation2025} and TabPFN \cite{hollmannAccuratePredictionsSmall2025} follow this approach, using structural causal models to generate diverse classification tasks.



\subsection{Research Questions}
Despite the centrality of pre-training data to TFM performance, surprisingly little is known about how these three corpora relate to one another.
We therefore structure this work around the following sequential research questions:
\begin{description}
    \item[RQ1: Do synthetic priors cover real tables?]
    Synthetic priors are designed using the causal processes that produce real tables, but whether they actually do is an open question. We show that synthetic priors occupy a very small region of the space occupied by real tables, but that the space of synthetic tables is well-covered by real tables.

    \item[RQ2: Do curated datasets cover web-scraped tables?]
    Curated datasets make use of tables known to be `clean' and useful for ML algorithms, and here we show that web-scraped and curated tables occupy broadly overlapping regions of feature space, though individual tables remain distinguishable by source. The distributional overlap explains why pre-training on comparatively few curated tables yields comparable performance.

    \item[RQ3: Can synthetic priors be optimised to cover real tables?]
    Given that neither web-scraped or curated datasets are well-covered by synthetic priors, the question remains as to whether the same prior design can be parameterised to match these real tables. Here we show that multiple methods of optimisation yield little improvement in coverage, with near perfect separation of synthetic and real tables after a highly extensive (86k trial) grid search.  

    \item[RQ4: Does proximity to synthetic prior impact performance?]
    Regardless of how well synthetic priors match real world distributions, an essential question is whether proximity to the prior actually impacts downstream performance.
    We present the preliminary finding that, under our feature sets, performance shows no clear correlation with proximity to the prior.

\end{description}

This paper addresses these questions directly, collecting archetypal corpora for each school of pre-training dataset.
We characterise each corpus using various feature sets capturing structural and distributional properties of individual tables, and measure sample and distribution-level similarities using discriminators and coverage metrics from the generative learning literature. 
Code is available online via GitHub (Supplemental for submission).

\section{Background and Related Work}
\label{sec:background}



Our focus is on the pre-training corpora used for Tabular Foundation Models (TFMs), how they compare to and cover each other, and whether this is relevant for downstream performance.

\subsection{Curated Datasets}

Curated datasets are assembled by curating tables from established machine learning repositories such as OpenML \cite{vanschorenOpenMLNetworkedScience2014} and the UCI Machine Learning Repository \cite{bayUCIKDDArchive2000} or the competition platform Kaggle.
Rather than collecting new data, this approach aggregates existing benchmarks into a unified corpus, applying filters for quality, schema consistency, and licence compatibility. TabFM \cite{wenSupervisedGenerativeNovel2024} is the primary example in this setting, comprising approximately 380 datasets drawn from Kaggle after deduplication and normalisation of column types.

A known limitation of curated corpora is that they inherit the biases of the repositories from which they are drawn. Certain domains, particularly medical and financial datasets, are heavily overrepresented, while the distribution of dataset sizes skews small, reflecting the historical tendency of benchmark repositories to favour datasets amenable to classical ML methods \cite{liuTalentTabularAnalytics2025}. These biases may limit the diversity of structural patterns seen during pre-training.

\subsection{Web-Scraped corpora}

An alternative to manual curation is to harvest tabular data at scale from the web. The T4 dataset \cite{gardnerLargeScaleTransfer2024a} and its respective TFM Tabula-8B follow this approach, extracting tables from HTML pages and other structured web sources to produce corpora orders of magnitude larger than hand-curated alternatives. The principal advantage is diversity: web-scraped corpora capture a wide range of domains, table sizes, and column types that are unlikely to appear in curated benchmark repositories.

This scale comes at a cost. Web-scraped tables are noisy, often containing malformed schemas, merged header rows, or semantically inconsistent columns. Licencing is also ambiguous, as the provenance of individual tables is difficult to trace \cite{gardnerLargeScaleTransfer2024a}. Pre-processing pipelines therefore play a critical role in determining the effective quality of the resulting corpus.

\subsection{Synthetic Priors}

Synthetic priors avoid real data entirely, instead sampling tables from a parametric generative prior. This approach offers several practical advantages: there are no licensing constraints, data can be generated at arbitrary scale, and the distribution can be controlled directly via the prior's hyperparameters. Early work in this vein includes the priors used in Prior-Fitted Networks (PFN) and TabPFN \cite{hollmannAccuratePredictionsSmall2025}, which generate synthetic classification tasks from simple Bayesian network structures.
An overview of priors --- and their notably limited public availability --- is available in the Appendix, Table~\ref{tab:synthetic-priors}.

TabICL \cite{quTabICLTabularFoundation2025} extends this idea using structural causal models (SCMs) as the generative mechanism. A SCM defines a directed acyclic graph over variables, with each node's value determined by a learned or sampled function of its parents plus noise. TabICL parameterises this process via a hyperparameter vector $\theta \in \Theta$ that governs choices such as the SCM type, the tree model family used to instantiate causal mechanisms, class balance, and the probability of generating categorical columns. Sampling from the prior thus yields synthetic tables whose statistical properties are determined by $\theta$.

The fundamental limitation of synthetic priors is distributional mismatch: there is no guarantee that tables sampled from the prior resemble real-world tabular data in their structural or distributional properties. Whether this mismatch is consequential for downstream ICL performance, and whether it can be reduced by tuning $\theta$, is the central question of this paper.

\section{Method}
\label{sec:method}

In this work we investigate, according to our research questions, the degree to which web-scraped, curated and synthetic prior tabular datasets cover the same spaces.
We conduct this through two primary metrics:
\begin{description}
    \item[Discriminators:] Samples that broadly cover the same space should be indistinguishable to a given model. Over the feature space described below, we fit discriminators to predict which class of dataset a table belongs to. This assesses the realism of samples on an individual basis, and is common in comparing sets of samples in generative modelling \cite{borjiProsConsGAN2019, goodfellowGenerativeAdversarialNets2014}.
    \item[Coverage:] A more direct distance-based metric, here we make use of neighbours-based coverage metrics \cite{kynkaanniemiImprovedPrecisionRecall2019, naeemReliableFidelityDiversity2020, sajjadiAssessingGenerativeModels2018} to assess the degree to which these tabular datasets overlap in a given feature space. These metrics are commonly applied in generative modelling, assessing how well a given model produces samples that cover the original data distribution.
\end{description}

We denote a table as $t$ with $|R|$ rows and $|C|$ columns. Column $j$ is written $C_j$, with cardinality $||C_j||$ denoting the number of unique values. The uniqueness ratio $\kappa_j = ||C_j||/|R|$ gives the proportion of unique values; we treat columns with $\kappa < 0.2$ as categorical. 

\subsection{Features}

Given the challenges of operating over datasets of tables (column and row order invariance, heterogenous column types, etc) it is necessary to move tables into a unified feature space.
We build a set of aggregate features that summarise table, column and inter-column patterns and distributions.
Features are, with the exception of column count, invariant to the number of columns and rows in a table, and in our later optimisation sections the column count becomes effectively redundant.
Details on our constructed features can be found in Appendix~\ref{sec:features}, and a discussion of their simplicity in Section~\ref{sec:feature:simplicity}.

It is worth noting that learnt embeddings and projections are not necessary for RQ1--3; it is only necessary to demonstrate a lack of coverage within a given feature space to answer these questions.
While learnt embeddings are likely more expressive, they would restrict our findings to coverage \textit{for a given model}, and as a result are not appropriate.
In particular, TabICL embeddings risk boot-strapping in these model-agnostic findings, though we do deploy them in Section~\ref{sec:rq4} as features maximally expressive \textit{with reference to TabICL and its prior distribution}.


We perform limited ablation over this feature set to ensure robustness in our results, using the following feature sets:

\begin{itemize}
    \item \textbf{Full} ($d = 70$): for table-level analysis, the complete feature vector $\phi(t)$.
    \item \textbf{Scalars} ($d = 9$): for table-level analysis, distributional scalar features, excluding the number of columns.
    \item \textbf{Histograms} ($d = 60$): for table-level analysis, average histogram mean and standard deviation features only.
    \item \textbf{Col. Hists} ($d=50$) for column-level analysis, cumulative column histograms extracted from individual tables.
    \item \textbf{Corr. Hists} ($d = 50$) for inter-column level analysis, correlation histograms over tables.
\end{itemize}

These ablations and varied feature sets allow us to validate the robustness of our findings, showing that our findings are neither specific to only one feature set, nor reliant on one single feature.

\subsection{Discriminator}

Given a set of real tables $\mathcal{T}_r$ and synthetic tables $\mathcal{T}_s$, we train a binary discriminator to distinguish between the two populations. Each table $t$ is embedded as $\phi(t) \in \mathbb{R}^{70}$ using the above-described features and assigned label $y = 1$ (real) or $y = 0$ (synthetic), yielding dataset
\begin{equation}
    \mathcal{D} = \{(\phi(t),\, 1) : t \in \mathcal{T}_r\} \cup \{(\phi(t),\, 0) : t \in \mathcal{T}_s\}.
\end{equation}

Given the number of necessary trials, concessions are made for speed and efficiency.
We use XGBoost classifiers trained on stratified splits of $\mathcal{D}$. 
A value of $\text{AUC} = 0.5$ indicates the discriminator performs at chance, with synthetic and real tables indistinguishable, while $\text{AUC} = 1.0$ indicates perfect separability. 
Thus $\text{AUC}$ is a proxy for \emph{prior realism}, and all prior optimisation methods minimise it.

\subsection{Coverage}


In addition to measuring  \textit{sample} realism, we use coverage as a measure of distributional overlap.
This is a set of methods and metrics developed primarily in generative modelling.
\citet{sajjadiAssessingGenerativeModels2018} first frame `recall' and `precision' for generative models as measures of distributional coverage.
\citet{kynkaanniemiImprovedPrecisionRecall2019} then extend this work using manifold estimation, with \citet{naeemReliableFidelityDiversity2020} proposing k-NN-based metrics along this theme which are  related to the metrics we describe below.

We measure the fraction of the real distribution not covered by the synthetic prior via a $k$-nearest-neighbour estimator, as well as the inverse, the fraction of the synthetic distribution not covered by the real.
Feature vectors are normalised to $[0, 1]$. 
A coverage threshold $\delta_{0.95}$ is set as the 95th percentile of mean within-synthetic $k$-NN distances at $k = 5$. 
A real table is considered \emph{uncovered} if its mean distance to its $k$ nearest
synthetic neighbours exceeds this threshold:
\begin{equation}
    f_{\text{unc}} = \frac{1}{|\mathcal{T}_r|}\sum_j \mathbf{1}\!\left[\bar{d}_k(r_j,\, \mathcal{T}_s) > \delta_{0.95}\right].
\end{equation}

\begin{description}
    \item[Coverage:] \textit{A point $p$ is \emph{covered} by a set $Q$, written $p \in Q$,
    if its mean distance to its $k$ nearest neighbours in $Q$ does not exceed $\delta_{0.95}$.
    For two sets $P$ and $Q$, we write $P \in Q$ to denote the proportion of points
    in $P$ covered by $Q$:}
    \begin{equation}
        P \in Q \;=\; \frac{1}{|P|}\sum_{p \in P} \mathbf{1}[p \in Q].
    \end{equation}

    \item[Recall:] \textit{The \emph{Recall} of $Q$ with respect to $P$ is the dataset-level
    coverage of $P$ by $Q$.}

    \item[Precision:] \textit{The \emph{Precision} of $Q$ with respect to $P$ is the dataset-level
    coverage of $Q$ by $P$.}
\end{description}

A crucial note on coverage is that, while qualitative findings hold from these metrics, quantitative results are comparable only under the same feature and distance metrics.
Reported recall and precision values should therefore be interpreted as lower and upper bounds respectively, rather than absolute characterisations of distributional overlap.





\begin{figure}[ht]
    \centering
    \includegraphics[width=0.8\linewidth]{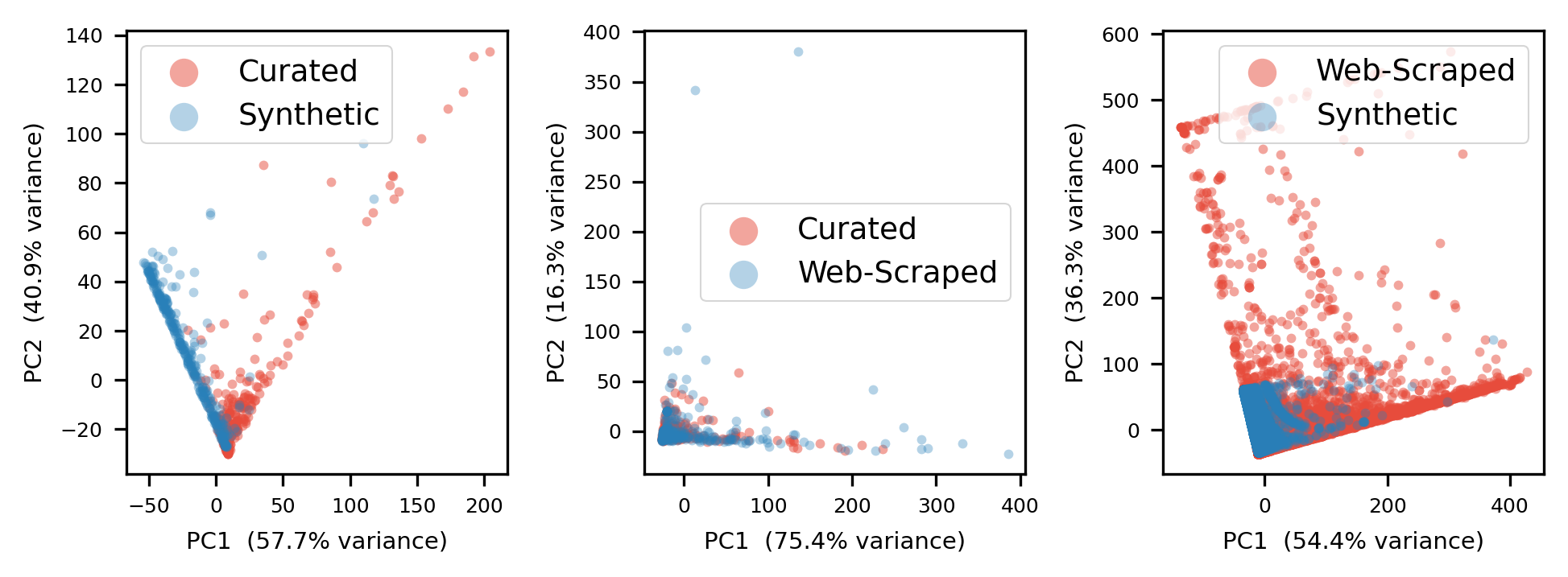}
    \caption{PCA projections of cumulative histograms for the Curated, Web-Scraped and Synthetic datasets within our full table-level feature space. Sample count is equal for each pair.}
    \label{fig:pairwise-pca}
\end{figure}

\subsection{Datasets}

Curated, web-scraped and synthetic prior-based datasets all aim to include enough statistical patterns that a TFM can downstream be applied to an arbitrary tabular dataset, the intent being that some of the downstream patterns will have appeared during pre-training.
To this extent, it is fair to say that the intent of these pre-training datasets is to \textit{cover} the space of downstream tables.

\textbf{Curated and Web-Scraped} We select representative examples of web-scraped and curated datasets of real-world tables.
As a web-scraped dataset we use the \textbf{T4} corpus \cite{gardnerLargeScaleTransfer2024a}.
This dataset was used to pre-train the model Tabula-8B, a language-model based TFM, and is a trimmed version of the massive TabLib \cite{eggert_tablib_2023} corpus, sourced from GitHub and CommonCrawl.
As a curated dataset, we use the collection of Kaggle datasets presented in \citet{wenSupervisedGenerativeNovel2024}, 384 mixed classification and regression datasets selected by various properties (e.g. row and column count, completeness).
This dataset was used to train the TFM TabFM; where appropriate we refer to the dataset by the moniker \textbf{FM}.

\textbf{Synthetic Prior} It remains to select a reasonable prior framework for generating tables.
We use that from TabICL \cite{quTabICLTabularFoundation2025}, which builds on the successes of the TabPFN \cite{hollmannAccuratePredictionsSmall2025} prior, and performs very strongly on benchmarks when compared with other single-prior methods.
We study a single prior, TabICL, for two reasons: first, it is the only canonical synthetic prior whose default parameters and generation code are fully open-source (TabPFN does not release prior parameters; Mitra does not distribute data-level code); second, restricting to one prior makes the 86k-configuration optimisation in Section \ref{sec:optimisation} computationally tractable.

A notable pattern in TFM development is closed-source code; Mitra \cite{zhangMitraMixedSynthetic2025} does not distribute data-level code, and TabPFN \cite{hollmannAccuratePredictionsSmall2025} do not share  prior parameters.
Comparisons without access to default parameters are simply impossible.
Despite this, TabICL functions very similarly  to TabPFN and other priors, including mixture models like Mitra, all of which rely on SCM-based modelling.
As a result, TabICL's priors are a best-case, with both strong, canonical performance and default parameters and code both open-source.
TabICL, within the literature of synthetic priors, builds on the successes of other priors, meaning that our findings may also carry over to these and other similar priors should their parameters become available.
We encourage future research to release open-source parameters so that this analysis can be extended.


\begin{table}[ht]
\caption{Bootstrap AUC and two-directional coverage across all experiments. The upper section shows unoptimised pairwise comparisons grouped by research question. The middle section shows the top-5 grid-search configurations (AUC-optimised) evaluated against T4. The lower section shows the best Bayesian-optimised configurations (joint AUC + coverage loss, Equation~\ref{eqn:triple-loss}) validated at larger sample sizes. Recall = A covered by B (A$\in$B); Precision = B covered by A (B$\in$A).}
\label{tab:combined-results}
\small
\centering
\begin{tabular}{p{0.05\linewidth} p{0.05\linewidth} p{0.13\linewidth} p{0.04\linewidth} p{0.22\linewidth} p{0.13\linewidth} p{0.13\linewidth}}
\toprule
\addlinespace[4pt]
\multicolumn{3}{l}{\textbf{Dataset}} &
&
\multicolumn{1}{l}{\textbf{Bootstrap AUC}} &
\multicolumn{2}{l}{\textbf{Coverage}} \\
\cmidrule(lr){1-4}  \cmidrule(lr){5-5} \cmidrule(lr){6-7}
\textbf{A} & \textbf{B} & \textbf{Features} & \multicolumn{1}{l}{\textbf{Rank}} &
\textbf{Mean $\pm$ Std} &
\textbf{Recall} & \textbf{Precision} \\
\midrule
\multicolumn{7}{l}{\textit{Validation: Self-Comparisons Between Subsets}} \\
\midrule
\multicolumn{2}{c}{T4} & \multirow{3}{*}{Full}        & & $0.498  \pm  0.015$ &   0.940   &   0.957 \\
\multicolumn{2}{c}{ICL} & & & $0.501  \pm  0.037$   & 0.944   &   0.963 \\
\multicolumn{2}{c}{FM} & & & $0.500  \pm  0.009$   & 0.938   &   0.955 \\
\midrule
\multicolumn{7}{l}{\textit{Unoptimised: Synthetic vs.\ Web-scraped (RQ1)}} \\
\midrule
\multirow{3}{*}{ICL} & \multirow{3}{*}{T4}
 & Full        & & $0.9997 \pm 0.0001$ & 0.0882 & 0.9880 \\
 && Histograms   & & $0.9985 \pm 0.0003$ & 0.1617 & 0.9849 \\
 && Scalars & & $0.9991 \pm 0.0002$ & 0.0854 & 0.9318 \\
 \cmidrule(lr){3-7}
 && Col. Hists & & $0.977 \pm 0.003$ & 0.250 & 0.987 \\
 && Corr. Hists && $0.988 \pm 0.011$ & 0.187 & 1.00 \\
\midrule
\multicolumn{7}{l}{\textit{Unoptimised: Synthetic vs.\ Curated (RQ1)}} \\
\midrule
\multirow{3}{*}{ICL} & \multirow{3}{*}{FM}
 & Full        & & $0.9972 \pm 0.0029$ & 0.1604 & 0.9780 \\
 && Histograms   & & $0.9943 \pm 0.0049$ & 0.2421 & 0.8742 \\
 && Scalars & & $0.9958 \pm 0.0037$ & 0.2736 & 0.6132 \\
\cmidrule(lr){3-7}
 && Col. Hists && $0.972 \pm 0.007$ & 0.421 & 0.970 \\
 && Corr. Hists && $0.983 \pm 0.014$ & 0.276 & 0.992 \\
\midrule
\multicolumn{7}{l}{\textit{Unoptimised: Web-scraped vs.\ Curated (RQ2)}} \\
\midrule
\multirow{3}{*}{T4} & \multirow{3}{*}{FM}
 & Full        & & $0.9133 \pm 0.0237$ & 0.9748 & 0.8302 \\
 && Histograms   & & $0.8884 \pm 0.0276$ & 0.9748 & 0.8836 \\
 && Scalars & & $0.8972 \pm 0.0250$ & 0.9906 & 0.8805 \\
\cmidrule(lr){3-7}
 && Col. Hists && $0.906 \pm 0.013$ & 0.968 & 0.852 \\
 && Corr. Hists && $0.841 \pm 0.052$ & 0.951 & 0.902 \\
\midrule
\multicolumn{7}{l}{\textit{Grid search optimised (AUC), ICL vs.\ T4 --- top-5 ranked configurations (RQ3)}} \\
\midrule
\multirow{3}{*}{ICL} & \multirow{3}{*}{T4} & \multirow{3}{*}{Full}
 & 1 & $0.9989 \pm 0.0013$ & 0.063 & 0.996 \\
 &&& 2 & $0.9987 \pm 0.0014$ & 0.080 & 0.992 \\
 &&& 3 & $0.9985 \pm 0.0016$ & 0.125 & 0.991 \\
\midrule
\multicolumn{7}{l}{\textit{Bayesian optimised (Equation~\ref{eqn:triple-loss}), ICL vs.\ real datasets --- best run per dataset (RQ3)}} \\
\midrule
\multirow{2}{*}{ICL} & FM & \multirow{2}{*}{Full} & 1 & $0.995 \pm 0.018$ & 0.194 & 0.981 \\
 & T4 && 1 & $0.997 \pm 0.006$ & 0.218 & 0.978 \\
\bottomrule
\end{tabular}

\end{table}

\section{Comparisons Between Datasets}
\label{sec:comparision}

In this section we address our first two research questions through treating datasets as-is, without optimisation towards either dataset of real tables.
In Figure~\ref{fig:pairwise-pca}, we show PCA projections of the feature sets for each dataset plotted against each other.
Figures showing cumulative coverage curves, feature importances and aggregates of our histogram features are available in the Appendix.
We present our results for discriminators and coverage in Table~\ref{tab:combined-results}, including validation of our discriminator and coverage metrics over two independent 1000-size samples of the respective dataset.
An ablation study for Recall and Precision is also available in Appendix~\ref{sec:ablation}.

\subsection{RQ1: Real vs. Synthetic}

Discriminator AUCs between ICL and our datasets of real tables are consistently high, within-margin of perfect performance when considering deviations during bootstrapping (200 re-samples). 
Coverage is similarly very low between ICL and our real tables, and comparatively high in reverse, indicating that the real tables cover the synthetic far more effectively than the inverse.

Feature importances indicate that histogram features prove the most useful for each pairwise discriminator (Figure~\ref{fig:pairwise-feature-importance}).
On comparing synthetic and real datasets, and with consideration of our visualisation in Figure~\ref{fig:pairwise-hist-features}, we note that at intermediate histogram bins ICL columns are consistently lower than columns from real tables, and have low variance within that region, making the middle range of column distributions highly informative for discriminators.
Discriminators remain highly performant when histogram features are included, indicating that simple summary statistics (in-particular the number of columns and the skewness of distributions) are still highly informative.

We also investigate the same metrics on column and inter-relation level distributions.
Over columns we use cumulative histograms, and over correlations simple histograms.
These results are reported in Table~\ref{tab:combined-results} and visualised through PCA in Figure~\ref{fig:pairwise-pca-columns}, with cumulative column histograms in Figure~\ref{fig:column-examples}.

Column and correlation AUCs are somewhat lower than for the whole feature set, due to decreased expressiveness compared to our full features which compare both.
Recall is also higher, though still comparably low, again due to the limited expressiveness of these feature sets.
That being said, AUCs are still near-perfect for Real/Synthetic discriminators, with recall consistently below 50\%.

From these results we can conclude that with default parameters, ICL priors produce tables with un-realistic patterns within columns and the distribution of synthetic tables is a narrow region of the space which real tables occupy.
Further, from column and correlation histograms as features, it is clear that these patterns hold true both within columns and in how columns relate to one another.

\subsection{RQ2: Curated vs. Web-Scraped}

Here we see significantly lower discriminator AUCs than for synthetic and real tables, though performance is still strong, indicating good sample-level discrimination.
Coverages are high in both directions, with T4 covering almost all of the FM tables, and FM covering the large majority ($>83\%$) of T4 tables.
While histogram features still prove the most informative, none of them contribute a gain of more than 6\%, indicating a more challenging classification task.

These results support two distinct conclusions. At the distributional level, the two corpora are near-interchangeable: high bidirectional coverage indicates that pre-training on either source encounters similar support in feature space, explaining why comparatively few curated datasets yield competitive performance. 
At the sample level, however, individual tables retain systematic source signatures (likely artefacts of curation filtering and web-scraping noise) sufficient for high discriminator AUCs.


\subsection{RQ3: Optimisation of Priors}
\label{sec:optimisation}


In this section we explore whether the priors can  be optimised to improve their poor coverage of real tables.
The TabICL generative prior is parameterised $\theta \in \Theta$ governing structural choices such as SCM type, tree model family, class balance, and the probability of generating categorical columns.
We seek
    $\theta^* = \operatorname*{argmin}_{\theta \in \Theta}\; \text{AUC}(\theta)$,
where $\text{AUC}(\theta)$ is the discriminator score when $N_{\text{eval}} = 200$ synthetic tables are drawn from the prior parameterised by $\theta$. 
The search space, i.e. the parameters for the TabICL prior over which we attempt to optimise, is outlined in the Appendix.
We make the trivial optimisation of sampling column counts from the distribution of columns counts in the respective real dataset, meaning that this can be effectively removed from our search space.

We attempt optimisation over AUC using Bayesian and Genetic optimisers (Appendix~\ref{app:optimisation}), yielding no significant improvement in AUCs with either strategy.
Following the lack of convergence from these optimisers, we perform an exhaustive grid-search over AUC.

\textbf{Grid Search - AUC.}
We construct a discrete grid by sampling each continuous or integer parameter at $G$ evenly-spaced values between its bounds, and taking the Cartesian product. Conditional parameters are only included where relevant, reducing the search space substantially. 
At our grid resolution of $G=6$, chosen to remain somewhat computationally tractable, this results in more than 86 thousand prior parameters and evaluations.
Even with only 200 real and synthetic samples, this is still highly computationally expensive.
This analysis was parallelised over a cluster of 6 nodes, each with 2 NVIDIA Grace CPU Superchips, and took a little less than 23 hours.

Figure~\ref{fig:grid-auc} shows AUCs over the whole grid search.
The lowest recorded AUC over this grid search is still $>0.95$, with the large majority not significantly different from the un-optimised synthetic prior.
Given the computational cost of this search we limit dataset sizes to 200 for both T4 and FM datasets, but perform subsequent analysis with top-ranked parameter combinations over 1000 samples to ensure robustness.
To avoid overfitting we reduce our discriminator trees to 100.

Following the grid search, the top-$K = 5$ configurations by AUC are subjected to a deeper post-hoc evaluation,  with 1000 synthetic tables sampled from the prior, compared against an equal sample of the T4 tables. Five analyses are run per configuration.

These results are presented in Table~\ref{tab:combined-results}.
Larger sample sizes lead to the resumption of near-perfect AUC, and coverage metrics are near-consistent with the unoptimised prior.
Taken together, this search has shown that the ICL prior simply cannot produce consistently realistic samples.
Given the lack of success in these optimisations we return to a Bayesian optimisation with a modified reward function that includes precision and recall, presented in Appendix~\ref{app:opt-triple}, but again find no significant improvement in coverage or AUC.

\begin{table}[ht]
\centering
\caption{Coverage and k-NN distance ($k=5$) correlation against coverage with varying relative performance metrics across the TALENT benchmark tables. $\downarrow$ where correlation indicates decreasing performance w.r.t. proximity to the ICL prior and $\uparrow$ for the inverse. No correlation results are statistically significant at $p \leq 0.05$.}
\label{tab:rank}
\small
\begin{tabular}{lrrrrrrrr}
\toprule
 & & & \multicolumn{2}{c}{Rank} & \multicolumn{2}{c}{$\frac{\textrm{AUC}_{\textrm{ICL}}}{\textrm{mean}(\textrm{AUC}_{\textrm{bench}})}$} & \multicolumn{2}{c}{$\frac{\textrm{AUC}_{\textrm{ICL}}}{\textrm{max}(\textrm{AUC}_{\textrm{bench}}})$} \\
\cmidrule(lr){4-5} \cmidrule(lr){6-7} \cmidrule(lr){8-9}
Features & Recall & Precision & $r$ & $p$ & $r$ & $p$ & $r$ & $p$ \\
\midrule
Full      & 0.171 & 0.988 & {$\downarrow \quad$ 0.084} & 0.283 & {$\uparrow \quad$ 0.036} & 0.646 & {$\uparrow \quad$ 0.013} & 0.872 \\
Column Hists.     & 0.244 & 0.973 & {$\downarrow \quad$ 0.103} & 0.188 & {$\downarrow \quad$-0.050} & 0.528 & {$\downarrow \quad$-0.041} & 0.602 \\
Corr. Hists.     & 0.308 & 0.995 & {$\downarrow \quad$ 0.145} & 0.172 & {$\uparrow \quad$ 0.002} & 0.982 & {$\uparrow \quad$ 0.045} & 0.674 \\ \midrule

TabICL Columns      & 0.226 & 0.783 & {$\downarrow \quad$ 0.125} & 0.111 & {$\uparrow \quad$ 0.021} & 0.787 & {$\downarrow \quad$-0.021} & 0.791 \\
TabICL Rows      & 0.659 & 0.841 & {$\downarrow \quad$ 0.138} & 0.078 & {$\downarrow \quad$-0.007} & 0.927 & {$\downarrow \quad$-0.066} & 0.401 \\
TabICL Concat. & 0.598 & 0.822 & {$\downarrow \quad$ 0.138} & 0.078 & {$\downarrow \quad$-0.002} & 0.979 & {$\downarrow \quad$-0.067} & 0.393 \\
\bottomrule
\end{tabular}
\end{table}

\section{Discussion}
\label{sec:feature-discussion}




Here we discuss two key points related to our above findings.
First, implications for performance downstream, and second, whether our sets are adequate for our findings.

\subsection{RQ4: Proximity to Prior and Performance}
\label{sec:rq4}

Despite the results presented above --- namely that the TabICL prior cannot be optimised to cover the space of real tables --- it remains to investigate whether this gap has a detectable relationship with downstream TFM performance.
In most machine-learning settings we would expect that increasing distance from the training distribution leads to degraded relative performance.
We test this expectation using both hand-crafted feature sets and, crucially, the model's own internal representations.

To this end, we assemble a collection of classical models (See Appendix~\ref{sec:classical-models}) along with TabICL.
We then apply this set of models on classification datasets from the more than 300 TALENT benchmarks \cite{liuTalentTabularAnalytics2025} and record the mean rank of ICL.
The TALENT benchmarks are specifically curated to reflect a large span of potential downstream applications, and are used in-literature to reflect the validity of an approach \cite{jiangRepresentationLearningTabular2026, quTabICLTabularFoundation2025}.
Having collected these rankings, we calculate 5-NN distance using \begin{enumerate*}[label=(\roman*), font=\textbf]
    \item Our full, table-level feature set
    \item Cumulative column histograms
    \item Correlation histograms over the table
    \item Internal embeddings from the pre-trained TabICL model
\end{enumerate*}.
We report quantitative results in Table~\ref{tab:rank}, and visualise in the Appendix.


The strongest evidence comes from the model's own internal representations (lower section of Table~\ref{tab:rank}).
TabICL's row and column embeddings are the learned features from which its predictions are directly derived; distance in this space is therefore the most functionally relevant measure of proximity to the prior.
Even here, correlations with performance remain below 0.14 and none approach statistical significance.
This result cannot be attributed to inadequacy of the feature representation --- the features \emph{are} the model's own view of the data, the same representation from which classification decisions are made.
If distance from the prior were degrading performance, we would expect it to be visible in the space the model itself uses.

The hand-crafted feature sets corroborate this finding across independent projections.
Under all six feature representations, across rank, mean-relative-AUC, and best-relative-AUC metrics, no correlation reaches significance at $p \leq 0.05$ (Table~\ref{tab:rank}).
With the 200 classification TALENT benchmarks, and at a power of $p \geq 0.8$ for significance at $p \leq 0.05$, we would detect correlations of $|r| > 0.2$. 
To further validate these results we perform the same analysis using the TabArena benchmark, which though consisting of fewer tables, does not overlap with TALENT.
These results are available in Appendix~\ref{sec:tabarena} --- the same null result holds.
We perform a partial correlation analysis controlling for dataset size, dimensionality, and class balance in Appendix~\ref{sec:confound}, finding that the null result holds across all three covariates.

Together with RQ1--RQ3, this presents a coherent picture: synthetic prior pre-training is substantially out-of-distribution with respect to real tables, yet this distance does not predict downstream performance in the model's own representation space.
We note that dataset-level covariates such as size, dimensionality, and class balance may independently affect both distance and difficulty; a partial-correlation analysis is available in the Appendix showing that this is not the case.

The mechanism underlying this robustness is not well-understood, and we outline two candidate interpretations.
First, in-context learning may be inherently more robust to pre-training distributional shift than standard supervised learning, with the inference-time conditioning on task-specific examples compensating for out-of-distribution pre-training.
Second, the inductive biases instilled by SCM-based pre-training (causal structure, tree-based mechanisms) may constitute a form of implicit regularisation that transfers broadly regardless of low-level distributional alignment.
Directly testing these accounts through pre-training under alternative or degraded priors is left to future work due to computational constraints.
We provide a brief for this future work in Appendix Section~\ref{sec:future-work}.

\subsection{Choice of Features}
\label{sec:feature:simplicity}

The feature sets used in RQ1–RQ3 are intentionally simple. 
For those questions this is sufficient: if tables are easily separable under a simple feature set, they remain separable under any more expressive one, and it is therefore not necessary to exhaust the space of possible representations to establish distributional mismatch. 
For RQ4 the situation differs. 
There, we do not rely solely on hand-crafted features but also measure proximity using TabICL's own internal row and column embeddings, the representations from which its classification decisions are directly derived. 
A richer external feature set could not be more relevant to the question of whether the distributional gap affects performance than the space the model itself uses to make predictions, and the null result in RQ4 is therefore not one that a more expressive feature set could plausibly overturn.



\section{Conclusion}

We have investigated the distributional relationships between the three principal classes of pre-training corpora for tabular foundation models. The archetypal TabICL prior occupies a narrow and unrealistic region of the space of real tables. Discriminators achieve near-perfect separation under all feature sets, coverage of real tables by the synthetic prior is consistently below 25\%, and extensive optimisation of prior hyperparameters yields no meaningful improvement — the lowest AUC across more than 86 thousand grid evaluations remains above 95\%. 
The distributional mismatch is therefore not a tuning problem but a structural limitation of this prior. Given the mixed-inclusion architecture shared across priors in this class (Table 3), we conjecture this limitation extends beyond TabICL, though direct verification awaits open-source release of comparable prior parameters.
Curated and web-scraped datasets, by contrast, occupy broadly overlapping regions of feature space despite remaining distinguishable at the individual-table level, providing a principled explanation for the empirical observation that comparatively few curated datasets can match web-scale performance \cite{gardnerLargeScaleTransfer2024a, wenSupervisedGenerativeNovel2024}.

We find no evidence that this distributional gap predicts downstream ICL performance, across two benchmarks, three performance metrics, and a partial correlation analysis controlling for dataset-level covariates. This result holds also in TabICL's own representation space, the setting in which a coverage-based account of generalisation would most plausibly predict a signal. That no such signal is observed is more consistent with an account based on the inductive biases instilled by SCM-based pre-training than with one based on distributional coverage, though we note this remains a conjecture. We note that this use of internal embeddings is appropriate only for RQ4, where the question concerns a specific model's behaviour; for RQ1–RQ3, where the question is whether the corpora differ in their intrinsic distributional properties, proximity must be assessed independently of any particular model's learned representation. Direct investigation is left to future work, and we encourage the release of generation code and default parameters as a norm in TFM development to make such comparisons possible.

\bibliographystyle{abbrvnat}
\bibliography{references}

@misc{zhangLimiXUnleashingStructuredData2025,
	title = {{LimiX}: {Unleashing} {Structured}-{Data} {Modeling} {Capability} for {Generalist} {Intelligence}},
	shorttitle = {{LimiX}},
	url = {https://arxiv.org/abs/2509.03505v2},
	language = {en},
	urldate = {2026-03-27},
	journal = {arXiv.org},
	author = {Zhang, Xingxuan and Ren, Gang and Yu, Han and Yuan, Hao and Wang, Hui and Li, Jiansheng and Wu, Jiayun and Mo, Lang and Mao, Li and Hao, Mingchao and Dai, Ningbo and Xu, Renzhe and Li, Shuyang and Zhang, Tianyang and He, Yue and Wang, Yuanrui and Zhang, Yunjia and Xu, Zijing and Li, Dongzhe and Gao, Fang and Zou, Hao and Liu, Jiandong and Liu, Jiashuo and Xu, Jiawei and Cheng, Kaijie and Li, Kehan and Zhou, Linjun and Li, Qing and Fan, Shaohua and Lin, Xiaoyu and Han, Xinyan and Li, Xuanyue and Lu, Yan and Xue, Yuan and Jiang, Yuanyuan and Wang, Zimu and Wang, Zhenlei and Cui, Peng},
	month = sep,
	year = {2025},
}

@inproceedings{helliDriftresilientTabPFNIncontext2024,
	address = {Red Hook, NY, USA},
	series = {{NIPS} '24},
	title = {Drift-resilient {TabPFN}: in-context learning temporal distribution shifts on tabular data},
	volume = {37},
	isbn = {979-8-3313-1438-5},
	shorttitle = {Drift-resilient {TabPFN}},
	urldate = {2026-03-27},
	booktitle = {Proceedings of the 38th {International} {Conference} on {Neural} {Information} {Processing} {Systems}},
	publisher = {Curran Associates Inc.},
	author = {Helli, Kai and Schnurr, David and Hollmann, Noah and Müller, Samuel and Hutter, Frank},
	month = dec,
	year = {2024},
	pages = {98742--98781},
}

@misc{breejenFinetunedInContextLearning2025,
	title = {Fine-tuned {In}-{Context} {Learning} {Transformers} are {Excellent} {Tabular} {Data} {Classifiers}},
	url = {http://arxiv.org/abs/2405.13396},
	doi = {10.48550/arXiv.2405.13396},
	urldate = {2026-03-27},
	publisher = {arXiv},
	author = {Breejen, Felix den and Bae, Sangmin and Cha, Stephen and Yun, Se-Young},
	month = jan,
	year = {2025},
	note = {arXiv:2405.13396 [cs]},
}

@misc{dorogushCatBoostGradientBoosting2018,
	title = {{CatBoost}: gradient boosting with categorical features support},
	shorttitle = {{CatBoost}},
	url = {http://arxiv.org/abs/1810.11363},
	doi = {10.48550/arXiv.1810.11363},
	urldate = {2026-03-24},
	publisher = {arXiv},
	author = {Dorogush, Anna Veronika and Ershov, Vasily and Gulin, Andrey},
	month = oct,
	year = {2018},
	note = {arXiv:1810.11363 [cs]},
}

@inproceedings{keLightGBMHighlyEfficient2017,
	address = {Red Hook, NY, USA},
	series = {{NIPS}'17},
	title = {{LightGBM}: a highly efficient gradient boosting decision tree},
	isbn = {978-1-5108-6096-4},
	shorttitle = {{LightGBM}},
	url = {https://dl.acm.org/doi/10.5555/3294996.3295074},
	urldate = {2026-03-24},
	booktitle = {Proceedings of the 31st {International} {Conference} on {Neural} {Information} {Processing} {Systems}},
	publisher = {Curran Associates Inc.},
	author = {Ke, Guolin and Meng, Qi and Finley, Thomas and Wang, Taifeng and Chen, Wei and Ma, Weidong and Ye, Qiwei and Liu, Tie-Yan},
	month = dec,
	year = {2017},
	pages = {3149--3157},
}

@inproceedings{chenXGBoostScalableTree2016,
	address = {New York, NY, USA},
	series = {{KDD} '16},
	title = {{XGBoost}: {A} {Scalable} {Tree} {Boosting} {System}},
	isbn = {978-1-4503-4232-2},
	shorttitle = {{XGBoost}},
	url = {https://dl.acm.org/doi/10.1145/2939672.2939785},
	doi = {10.1145/2939672.2939785},
	urldate = {2026-03-24},
	booktitle = {Proceedings of the 22nd {ACM} {SIGKDD} {International} {Conference} on {Knowledge} {Discovery} and {Data} {Mining}},
	publisher = {Association for Computing Machinery},
	author = {Chen, Tianqi and Guestrin, Carlos},
	month = aug,
	year = {2016},
	pages = {785--794},
}

@misc{ericksonAutoGluonTabularRobustAccurate2020,
	title = {{AutoGluon}-{Tabular}: {Robust} and {Accurate} {AutoML} for {Structured} {Data}},
	shorttitle = {{AutoGluon}-{Tabular}},
	url = {http://arxiv.org/abs/2003.06505},
	doi = {10.48550/arXiv.2003.06505},
	urldate = {2026-03-24},
	publisher = {arXiv},
	author = {Erickson, Nick and Mueller, Jonas and Shirkov, Alexander and Zhang, Hang and Larroy, Pedro and Li, Mu and Smola, Alexander},
	month = mar,
	year = {2020},
	note = {arXiv:2003.06505 [stat]},
}

@article{cortesSupportvectorNetworks1995,
	title = {Support-vector networks},
	volume = {20},
	issn = {1573-0565},
	url = {https://doi.org/10.1007/BF00994018},
	doi = {10.1007/BF00994018},
	language = {en},
	number = {3},
	urldate = {2026-03-24},
	journal = {Machine Learning},
	author = {Cortes, Corinna and Vapnik, Vladimir},
	month = sep,
	year = {1995},
	pages = {273--297},
}

@article{breimanRandomForests2001,
	title = {Random {Forests}},
	volume = {45},
	issn = {1573-0565},
	url = {https://doi.org/10.1023/A:1010933404324},
	doi = {10.1023/A:1010933404324},
	language = {en},
	number = {1},
	urldate = {2026-03-24},
	journal = {Machine Learning},
	author = {Breiman, Leo},
	month = oct,
	year = {2001},
	pages = {5--32},
}

@inproceedings{sajjadiAssessingGenerativeModels2018,
	title = {Assessing {Generative} {Models} via {Precision} and {Recall}},
	volume = {31},
	url = {https://proceedings.neurips.cc/paper/2018/hash/f7696a9b362ac5a51c3dc8f098b73923-Abstract.html},
	urldate = {2026-03-24},
	booktitle = {Advances in {Neural} {Information} {Processing} {Systems}},
	publisher = {Curran Associates, Inc.},
	author = {Sajjadi, Mehdi S. M. and Bachem, Olivier and Lucic, Mario and Bousquet, Olivier and Gelly, Sylvain},
	year = {2018},
}

@inproceedings{naeemReliableFidelityDiversity2020,
	title = {Reliable {Fidelity} and {Diversity} {Metrics} for {Generative} {Models}},
	issn = {2640-3498},
	url = {https://proceedings.mlr.press/v119/naeem20a.html},
	language = {en},
	urldate = {2026-03-24},
	booktitle = {Proceedings of the 37th {International} {Conference} on {Machine} {Learning}},
	publisher = {PMLR},
	author = {Naeem, Muhammad Ferjad and Oh, Seong Joon and Uh, Youngjung and Choi, Yunjey and Yoo, Jaejun},
	month = nov,
	year = {2020},
	pages = {7176--7185},
}

@inproceedings{kynkaanniemiImprovedPrecisionRecall2019,
	title = {Improved {Precision} and {Recall} {Metric} for {Assessing} {Generative} {Models}},
	volume = {32},
	url = {https://proceedings.neurips.cc/paper/2019/hash/0234c510bc6d908b28c70ff313743079-Abstract.html},
	urldate = {2026-03-24},
	booktitle = {Advances in {Neural} {Information} {Processing} {Systems}},
	publisher = {Curran Associates, Inc.},
	author = {Kynkäänniemi, Tuomas and Karras, Tero and Laine, Samuli and Lehtinen, Jaakko and Aila, Timo},
	year = {2019},
}

@article{borjiProsConsGAN2019,
	title = {Pros and cons of {GAN} evaluation measures},
	volume = {179},
	issn = {1077-3142},
	url = {https://www.sciencedirect.com/science/article/pii/S1077314218304272},
	doi = {10.1016/j.cviu.2018.10.009},
	urldate = {2026-03-24},
	journal = {Computer Vision and Image Understanding},
	author = {Borji, Ali},
	month = feb,
	year = {2019},
	pages = {41--65},
}

@inproceedings{goodfellowGenerativeAdversarialNets2014,
	title = {Generative {Adversarial} {Nets}},
	volume = {27},
	url = {https://proceedings.neurips.cc/paper/2014/hash/f033ed80deb0234979a61f95710dbe25-Abstract.html},
	urldate = {2026-03-24},
	booktitle = {Advances in {Neural} {Information} {Processing} {Systems}},
	publisher = {Curran Associates, Inc.},
	author = {Goodfellow, Ian J. and Pouget-Abadie, Jean and Mirza, Mehdi and Xu, Bing and Warde-Farley, David and Ozair, Sherjil and Courville, Aaron and Bengio, Yoshua},
	year = {2014},
}

@article{liuTalentTabularAnalytics2025,
	title = {Talent: {A} {Tabular} {Analytics} and {Learning} {Toolbox}},
	volume = {26},
	issn = {1533-7928},
	shorttitle = {Talent},
	url = {http://jmlr.org/papers/v26/25-0512.html},
	number = {226},
	urldate = {2026-03-24},
	journal = {Journal of Machine Learning Research},
	author = {Liu, Si-Yang and Cai, Hao-Run and Zhou, Qi-Le and Yin, Huai-Hong and Zhou, Tao and Jiang, Jun-Peng and Ye, Han-Jia},
	year = {2025},
	pages = {1--16},
}

@article{vanschorenOpenMLNetworkedScience2014,
	title = {{OpenML}: networked science in machine learning},
	volume = {15},
	issn = {1931-0145},
	shorttitle = {{OpenML}},
	url = {https://dl.acm.org/doi/10.1145/2641190.2641198},
	doi = {10.1145/2641190.2641198},
	number = {2},
	urldate = {2026-03-24},
	journal = {SIGKDD Explor. Newsl.},
	author = {Vanschoren, Joaquin and van Rijn, Jan N. and Bischl, Bernd and Torgo, Luis},
	month = jun,
	year = {2014},
	pages = {49--60},
}

@article{bayUCIKDDArchive2000,
	title = {The {UCI} {KDD} archive of large data sets for data mining research and experimentation},
	volume = {2},
	issn = {1931-0145},
	url = {https://dl.acm.org/doi/10.1145/380995.381030},
	doi = {10.1145/380995.381030},
	number = {2},
	urldate = {2026-03-24},
	journal = {SIGKDD Explor. Newsl.},
	author = {Bay, Stephen D. and Kibler, Dennis and Pazzani, Michael J. and Smyth, Padhraic},
	month = dec,
	year = {2000},
	pages = {81--85},
}

@inproceedings{gardnerLargeScaleTransfer2024a,
	title = {Large {Scale} {Transfer} {Learning} for {Tabular} {Data} via {Language} {Modeling}},
	volume = {37},
	url = {https://proceedings.neurips.cc/paper_files/paper/2024/hash/4fd5cfd2e31bebbccfa5ffa354c04bdc-Abstract-Conference.html},
	doi = {10.52202/079017-1435},
	language = {en},
	urldate = {2026-03-24},
	booktitle = {Advances in {Neural} {Information} {Processing} {Systems}},
	author = {Gardner, Josh and Perdomo, Juan C. and Schmidt, Ludwig},
	month = dec,
	year = {2024},
	pages = {45155--45205},
}

@inproceedings{wenSupervisedGenerativeNovel2024,
	address = {New York, NY, USA},
	series = {{KDD} '24},
	title = {From {Supervised} to {Generative}: {A} {Novel} {Paradigm} for {Tabular} {Deep} {Learning} with {Large} {Language} {Models}},
	isbn = {979-8-4007-0490-1},
	shorttitle = {From {Supervised} to {Generative}},
	url = {https://dl.acm.org/doi/10.1145/3637528.3671975},
	doi = {10.1145/3637528.3671975},
	urldate = {2026-03-24},
	booktitle = {Proceedings of the 30th {ACM} {SIGKDD} {Conference} on {Knowledge} {Discovery} and {Data} {Mining}},
	publisher = {Association for Computing Machinery},
	author = {Wen, Xumeng and Zhang, Han and Zheng, Shun and Xu, Wei and Bian, Jiang},
	month = aug,
	year = {2024},
	pages = {3323--3333},
}

@article{shwartz-zivTabularDataDeep2022,
	title = {Tabular data: {Deep} learning is not all you need},
	volume = {81},
	issn = {1566-2535},
	shorttitle = {Tabular data},
	url = {https://www.sciencedirect.com/science/article/pii/S1566253521002360},
	doi = {10.1016/j.inffus.2021.11.011},
	urldate = {2026-03-24},
	journal = {Information Fusion},
	author = {Shwartz-Ziv, Ravid and Armon, Amitai},
	month = may,
	year = {2022},
	pages = {84--90},
}

@article{jiangRepresentationLearningTabular2026,
	title = {Representation {Learning} for {Tabular} {Data}: {A} {Comprehensive} {Survey}},
	issn = {1939-3539},
	shorttitle = {Representation {Learning} for {Tabular} {Data}},
	url = {https://ieeexplore.ieee.org/abstract/document/11369258},
	doi = {10.1109/TPAMI.2026.3657217},
	urldate = {2026-03-24},
	journal = {IEEE Transactions on Pattern Analysis and Machine Intelligence},
	author = {Jiang, Jun-Peng and Liu, Si-Yang and Cai, Hao-Run and Zhou, Qi-Le and Ye, Han-Jia},
	year = {2026},
	pages = {1--20},
}

@article{quinnLiteratureReviewExplainable2024,
	title = {Literature {Review} of {Explainable} {Tabular} {Data} {Analysis}},
	volume = {13},
	copyright = {http://creativecommons.org/licenses/by/3.0/},
	issn = {2079-9292},
	url = {https://www.mdpi.com/2079-9292/13/19/3806},
	doi = {10.3390/electronics13193806},
	language = {en},
	number = {19},
	urldate = {2026-03-24},
	journal = {Electronics},
	publisher = {Multidisciplinary Digital Publishing Institute},
	author = {Quinn, Helen O’Brien and Sedky, Mohamed and Francis, Janet and Streeton, Michael},
	month = sep,
	year = {2024},
}

@inproceedings{akoglu_rtm_2008,
	title = {{RTM}: {Laws} and a recursive generator for weighted time-evolving graphs},
	isbn = {978-0-7695-3502-9},
	doi = {10.1109/ICDM.2008.123},
	booktitle = {Proceedings - {IEEE} international conference on data mining, {ICDM}},
	author = {Akoglu, Leman and McGlohon, Mary and Faloutsos, Christos},
	year = {2008},
	pages = {701--706},
}

@inproceedings{akiba_optuna_2019,
	address = {New York, NY, USA},
	series = {{KDD} '19},
	title = {Optuna: {A} {Next}-generation {Hyperparameter} {Optimization} {Framework}},
	isbn = {978-1-4503-6201-6},
	shorttitle = {Optuna},
	url = {https://dl.acm.org/doi/10.1145/3292500.3330701},
	doi = {10.1145/3292500.3330701},
	urldate = {2026-03-13},
	booktitle = {Proceedings of the 25th {ACM} {SIGKDD} {International} {Conference} on {Knowledge} {Discovery} \& {Data} {Mining}},
	publisher = {Association for Computing Machinery},
	author = {Akiba, Takuya and Sano, Shotaro and Yanase, Toshihiko and Ohta, Takeru and Koyama, Masanori},
	month = jul,
	year = {2019},
	pages = {2623--2631},
}

@misc{zhangMitraMixedSynthetic2025,
	title = {Mitra: {Mixed} {Synthetic} {Priors} for {Enhancing} {Tabular} {Foundation} {Models}},
	shorttitle = {Mitra},
	url = {http://arxiv.org/abs/2510.21204},
	doi = {10.48550/arXiv.2510.21204},
	urldate = {2026-03-13},
	publisher = {arXiv},
	author = {Zhang, Xiyuan and Maddix, Danielle C. and Yin, Junming and Erickson, Nick and Ansari, Abdul Fatir and Han, Boran and Zhang, Shuai and Akoglu, Leman and Faloutsos, Christos and Mahoney, Michael W. and Hu, Cuixiong and Rangwala, Huzefa and Karypis, George and Wang, Bernie},
	month = oct,
	year = {2025},
	note = {arXiv:2510.21204 [cs]},
}

@inproceedings{quTabICLTabularFoundation2025,
	title = {{TabICL}: {A} {Tabular} {Foundation} {Model} for {In}-{Context} {Learning} on {Large} {Data}},
	issn = {2640-3498},
	shorttitle = {{TabICL}},
	url = {https://proceedings.mlr.press/v267/qu25d.html},
	language = {en},
	urldate = {2026-03-13},
	booktitle = {Proceedings of the 42nd {International} {Conference} on {Machine} {Learning}},
	publisher = {PMLR},
	author = {Qu, Jingang and Holzmüller, David and Varoquaux, Gaël and Morvan, Marine Le},
	month = oct,
	year = {2025},
	pages = {50817--50847},
}

@misc{eggert_tablib_2023,
	title = {{TabLib}: {A} {Dataset} of {627M} {Tables} with {Context}},
	shorttitle = {{TabLib}},
	url = {http://arxiv.org/abs/2310.07875},
	doi = {10.48550/arXiv.2310.07875},
	urldate = {2026-03-13},
	publisher = {arXiv},
	author = {Eggert, Gus and Huo, Kevin and Biven, Mike and Waugh, Justin},
	month = oct,
	year = {2023},
	note = {arXiv:2310.07875 [cs]},
}

@article{hollmannAccuratePredictionsSmall2025,
	title = {Accurate predictions on small data with a tabular foundation model},
	volume = {637},
	copyright = {2025 The Author(s)},
	issn = {1476-4687},
	url = {https://www.nature.com/articles/s41586-024-08328-6},
	doi = {10.1038/s41586-024-08328-6},
	language = {en},
	number = {8045},
	urldate = {2025-03-31},
	journal = {Nature},
	author = {Hollmann, Noah and Müller, Samuel and Purucker, Lennart and Krishnakumar, Arjun and Körfer, Max and Hoo, Shi Bin and Schirrmeister, Robin Tibor and Hutter, Frank},
	month = jan,
	year = {2025},
	note = {Publisher: Nature Publishing Group},
	pages = {319--326},
}

\clearpage
\newpage

\appendix

\section{Synthetic Priors}

\begin{table}[ht]
\centering
\small
\caption{Synthetic priors for tabular foundation model pre-training, ordered by paper date.
``Prior code'' means the data generation pipeline is publicly available, not merely inference code or trained weights.
``Covers / extends'' describes inclusion at the level of the prior's generative mechanism.}
\label{tab:synthetic-priors}
\setlength{\tabcolsep}{5pt}
\begin{tabular}{p{0.12\linewidth}  p{0.07\linewidth} c c p{0.14\linewidth} p{0.375\linewidth}}
\toprule
\textbf{Model}  & \textbf{Prior Family} & \textbf{Code} & \textbf{Params} & \textbf{Paper} & \textbf{Covers / extends} \\
\midrule

TabPFN v1
& SCM + BNN
& \xmark
& \xmark
& \citet{hollmannAccuratePredictionsSmall2025}, published 2025
& Root of the SCM lineage. Equal mixture of random-DAG SCMs (MLP edges) and Bayesian neural nets. The BNN component was not carried forward by any successor. \\\midrule


TabForestPFN
& SCM + Forest
& \cmark
& \cmark
& \citet{breejenFinetunedInContextLearning2025}, pre-print 2025 \textit{(arXiv)}
& Strictly \emph{includes} the v1 SCM prior as one component. The forest generator (decision-boundary-oriented) is an independent primitive. Direct ancestor of TabICL v1's and Mitra's tree-mixing strategies. \\\midrule


Drift-Resilient TabPFN
& Temporal SCM
& \cmark
& \xmark
& \citet{helliDriftresilientTabPFNIncontext2024}, published 2024
& Strictly \emph{includes} v1 SCM as its inner prior; wraps it in a second-level process that modulates DAG parameters over time. Does not incorporate forest or tree-ensemble components. \\\midrule


TabPFN v2 / v2.5
& SCM (enriched)
& \xmark
& \xmark
& \citet{hollmannAccuratePredictionsSmall2025}, published 2025
& Parallel refinement of v1's SCM, not a strict superset. DAGs grown via redirection sampling; edge functions include small NNs and decision trees. BNN component dropped. Prior generation explicitly not released. \\\midrule


TabICL v1/v2
& SCM + Tree-SCM
& \cmark
& \cmark
& \citet{quTabICLTabularFoundation2025}, published 2025
& Subsumes the v1 SCM prior (with extended activations) and adds a tree-based SCM branch (XGBoost models at DAG edges) inspired by TabForestPFN. Covers TabForestPFN's intent; implementation differs. \\\midrule


LimiX (16M / 2M)
& Hierarchical SCM
& \xmark
& \xmark
& \citet{zhangLimiXUnleashingStructuredData2025}, published 2025 \textit{(arXiv)}
& Parallel SCM lineage; hierarchical DAG generation with graph-aware and solvability-aware sampling. Does not include forest or tree-ensemble components. Broadest task scope (imputation, causal inference). \\\midrule


Mitra
& SCM + GBDT + RF + DT
& \xmark
& \xmark
& \citet{zhangMitraMixedSynthetic2025}, published 2025
& Broadest mixture prior. \emph{Includes} an SCM component and three tree-based generators (gradient boosting, random forest, decision tree) selected via a Generalizability Matrix (diversity $\times$ performance). \\

\bottomrule
\end{tabular}
\end{table}

Tabular foundation models (TFMs) trained via prior-data fitted networks (PFNs) approximate Bayesian inference over a hand-designed generative prior rather than learning from real-world datasets.
This makes the choice of synthetic prior the central design decision in TFM development: the prior encodes all inductive biases the model will bring to unseen tabular tasks, and its coverage of realistic data-generating processes directly determines downstream generalisation. 
Despite this centrality, the prior design space has received comparatively little systematic attention relative to architectural innovations, and crucially, the generation code is rarely released — making the prior something closer to proprietary IP than a scientific artefact.

Table~\ref{tab:synthetic-priors} surveys the synthetic priors used in the main PFN-style TFMs to date, ordered chronologically.
The dominant primitive is the structural causal model (SCM), in which a directed acyclic graph encodes causal dependencies between features and targets, with each edge function drawn from a parameterised family. 
TabPFN v1 \cite{hollmannAccuratePredictionsSmall2025} established this foundation, mixing SCM-generated datasets equally with Bayesian neural network samples; every successor dropped the BNN component and focused on enriching the SCM side.
Two parallel directions have emerged from this base.
The first refines DAG construction — TabPFN v2 \cite{hollmannAccuratePredictionsSmall2025} replaces fully-connected MLP edges with a redirection-sampling growth procedure, while LimiX \cite{zhangLimiXUnleashingStructuredData2025} introduces a hierarchical generation paradigm with explicit solvability control.
The second direction mixes the SCM prior with tree-based generators, motivated by the observation that gradient-boosted decision trees carry inductive biases that purely causal models lack: TabForestPFN \cite{breejenFinetunedInContextLearning2025} introduced this mixture, and both TabICL v1 \cite{quTabICLTabularFoundation2025} and Mitra \cite{zhangMitraMixedSynthetic2025} extended it, with Mitra offering the most principled treatment by selecting mixture components via a formal diversity-and-performance criterion.

Notably, only three of the seven priors surveyed (TabForestPFN \cite{breejenFinetunedInContextLearning2025}, Drift-Resilient TabPFN \cite{helliDriftresilientTabPFNIncontext2024}, and TabICL v1 \cite{quTabICLTabularFoundation2025}) have released their generation code; the remainder describe their priors in varying levels of paper detail but treat the generation pipeline as closed.
Only TabICL and TabForestPFN release default parameters for their own training runs; we opt for TabICL due to its popularity and publication status.

\clearpage
\newpage

\section{Comparisons Between Datasets}

\begin{figure}[ht]
    \includegraphics[width=\linewidth]{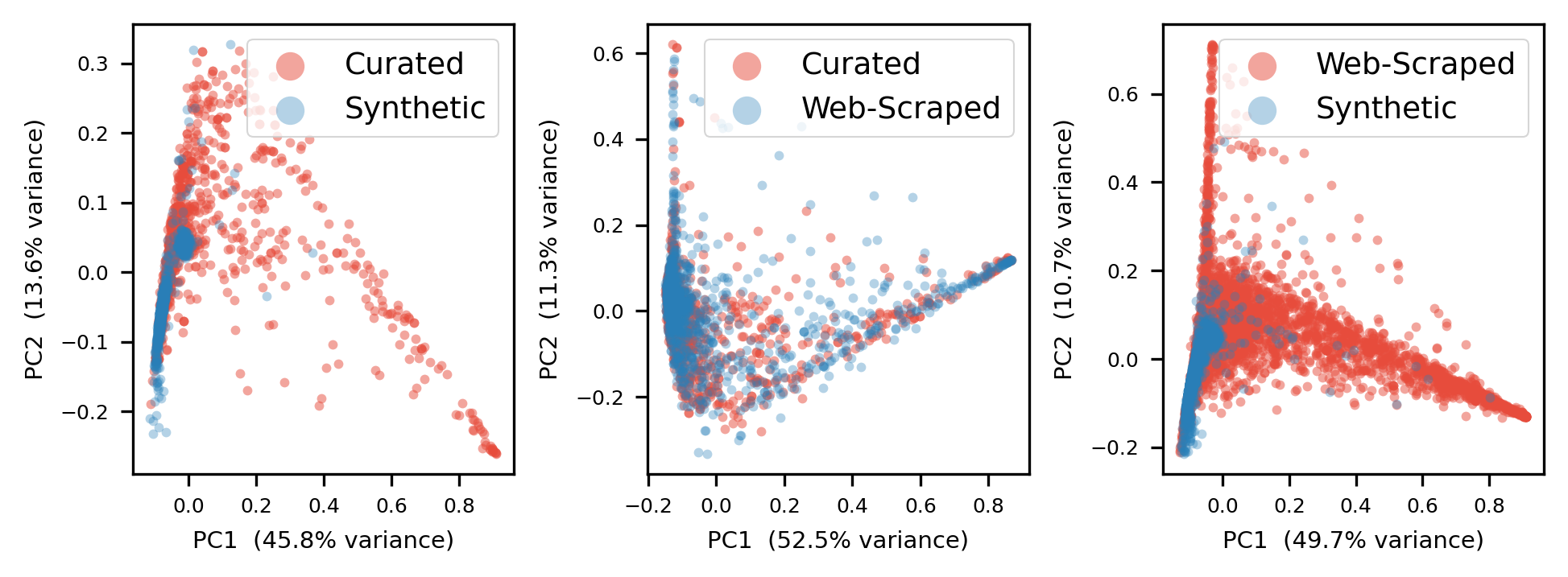}
    \caption{PCA projections of cumulative histograms for the curated FM dataset, the web-scraped T4 dataset, and the synthetic ICL datasets within with cumulative column histograms as a feature space. Sample count is equal for each pair. }
    \label{fig:pairwise-pca-columns}
\end{figure}

\subsection{Features}
\label{sec:features}

We begin with simple aggregate features over table structure:
\begin{description}
    \item[Num Cols:] The dimensionalities $|R|$ and $|C|$ of the table

    \item[Categorical Ratio] The proportion of columns with $\kappa < 0.2$, inferred as categorical under the uniqueness heuristic.
    \begin{equation}
        \frac{1}{|C|}\sum_{j=1}^{|C|} \mathbf{1}\!\left[\kappa_j < 0.2\right].
    \end{equation}
\end{description}

\textbf{Within-Column}
We then compute a set of features to express properties of the numeric or categorical values within columns, which are then aggregated over the table.
Let $\mathcal{C}_{\text{num}} = \{j : \kappa_j \geq 0.2\}$ denote the set of inferred numeric columns, with per-column mean $\mu_j = \mathbb{E}[C_j]$ and standard deviation $\sigma_j$.

\begin{description}
    \item[Mean cardinality for categorical columns:] Let $\mathcal{C}_{\text{cat}} = \{j : \kappa_j < 0.2\}$. The mean number of distinct values across inferred categorical columns, $\bar{\mathcal{C}_{\text{cat}}}$.
    \begin{equation}
        \frac{1}{|\mathcal{C}_{\text{cat}}|} \sum_{j \in \mathcal{C}_{\text{cat}}} \|C_j\|.
    \end{equation}

    \item[Max cardinality for categorical columns:] The maximum number of distinct values across inferred categorical columns.

    \item[Skewness:] The third standardised moment of each numeric column, averaged across all columns. 
    We summarise across columns via the mean $\overline{\text{skew}} = \frac{1}{|\mathcal{C}_{\text{num}}|}\sum_j \text{skew}(C_j)$ and standard deviation $\sigma_{\text{skew}}$.

    \item[Kurtosis:] The fourth standardised moment (excess), averaged across all columns.

    \item[Entropy for categorical columns:] The Shannon entropy of the empirical value distribution of each inferred categorical column.
\end{description}

\textbf{Between-Column}

Also necessary to express is how columns relate to one another; this inter-column structure is captured via the Pearson correlation matrix over numeric columns. Let $\mathbf{P} \in \mathbb{R}^{|\mathcal{C}_{\text{num}}| \times |\mathcal{C}_{\text{num}}|}$ be the matrix of pairwise Pearson correlations, and let $\mathcal{P} = \{|P_{jk}| : j \neq k\}$ denote the multiset of off-diagonal absolute correlations. 

We extract three summary statistics; the mean absolute correlation $\bar{\rho} = (1/|\mathcal{P})  | \cdot \sum_{r \in \mathcal{P}} r$, the standard deviation of absolute correlations $\sigma_\rho$, and the proportion of highly correlated pairs $r \geq 0.7$.

\subsection{Ablation - Recall, Coverage}
\label{sec:ablation}

\begin{figure}[ht]
    \centering
    \includegraphics[width=\linewidth]{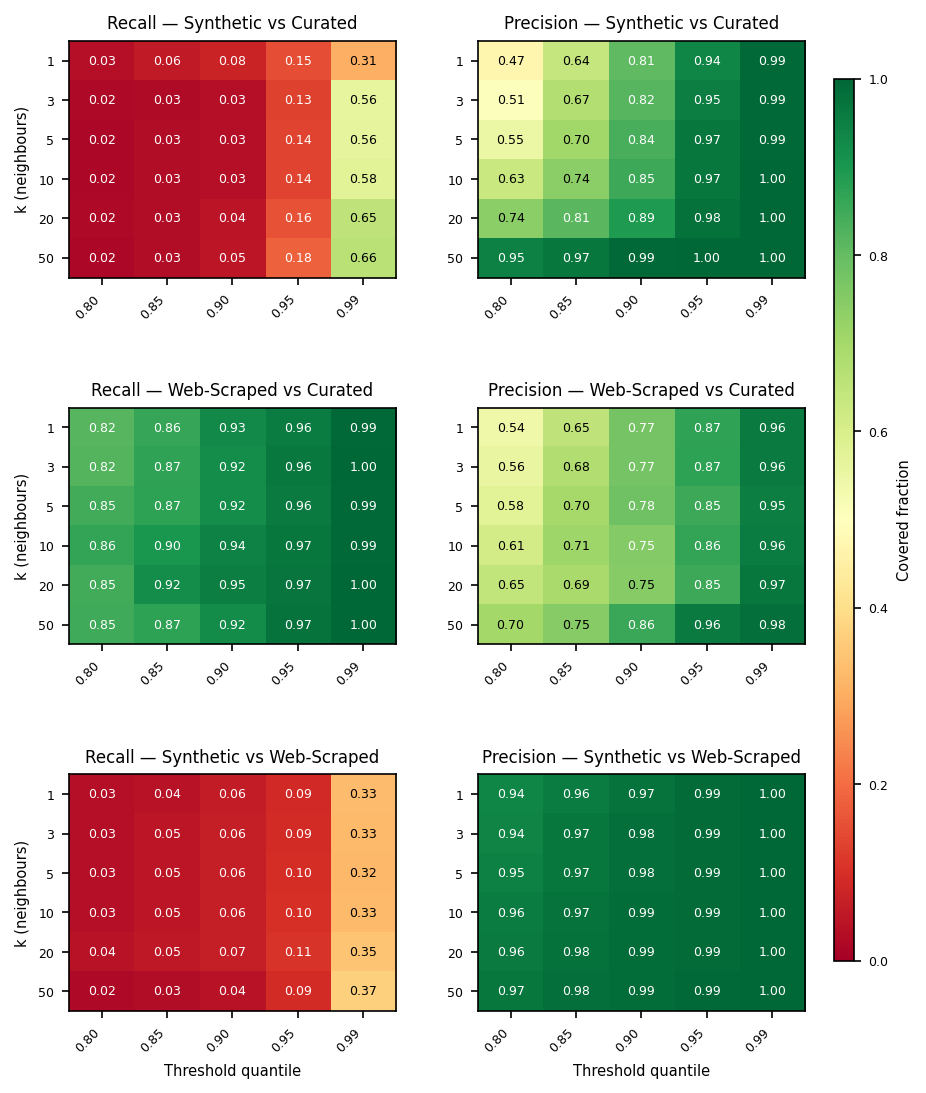}
    \caption{Our ablation study over the number of neighbours $k$ and threshold $\delta_k$ for recall and precision for each pairwise comparison of our default-parameter datasets.}
    \label{fig:ablation}
\end{figure}

We perform a limited ablation over the number of neighbours $k$ and the distance threshold $\bar{\delta}_k$ for each pairwise comparison between our datasets.
We use equal sized samples for each dataset comparison, meaning that T4 and ICL are limited to 318 tables when comparing to FM, and 5k samples when compared to one another.
We present the results, through annotated heatmaps, in Figure~\ref{fig:ablation}.

\begin{figure}[ht]
    \centering
    \includegraphics[width=\linewidth]{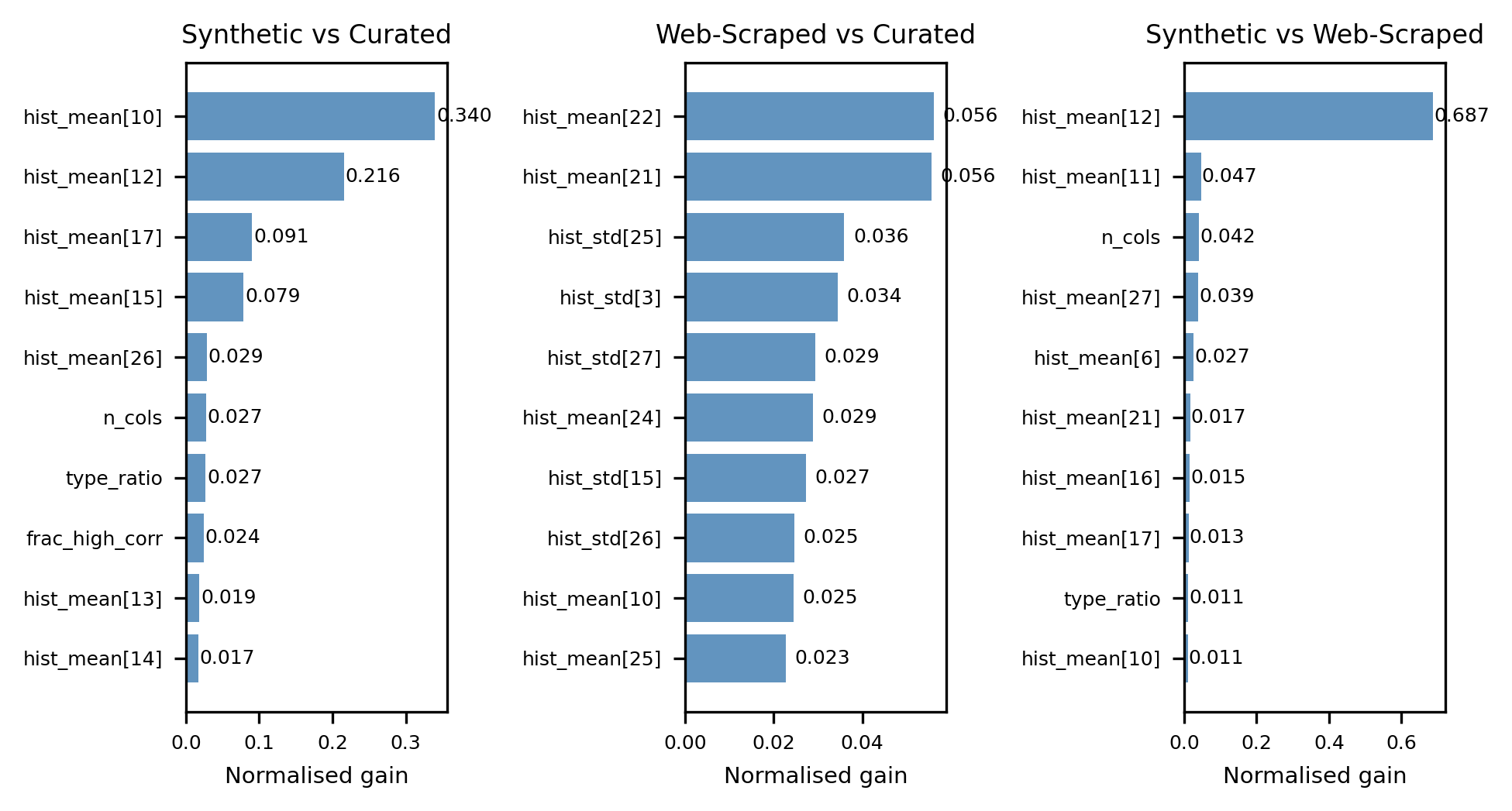}
    \caption{Feature importances, assessed from performance gains through tree splits, for each pairwise discriminator over our three datasets.}
    \label{fig:pairwise-feature-importance}
\end{figure}

\begin{figure}[ht]
    \centering
    \includegraphics[width=\linewidth]{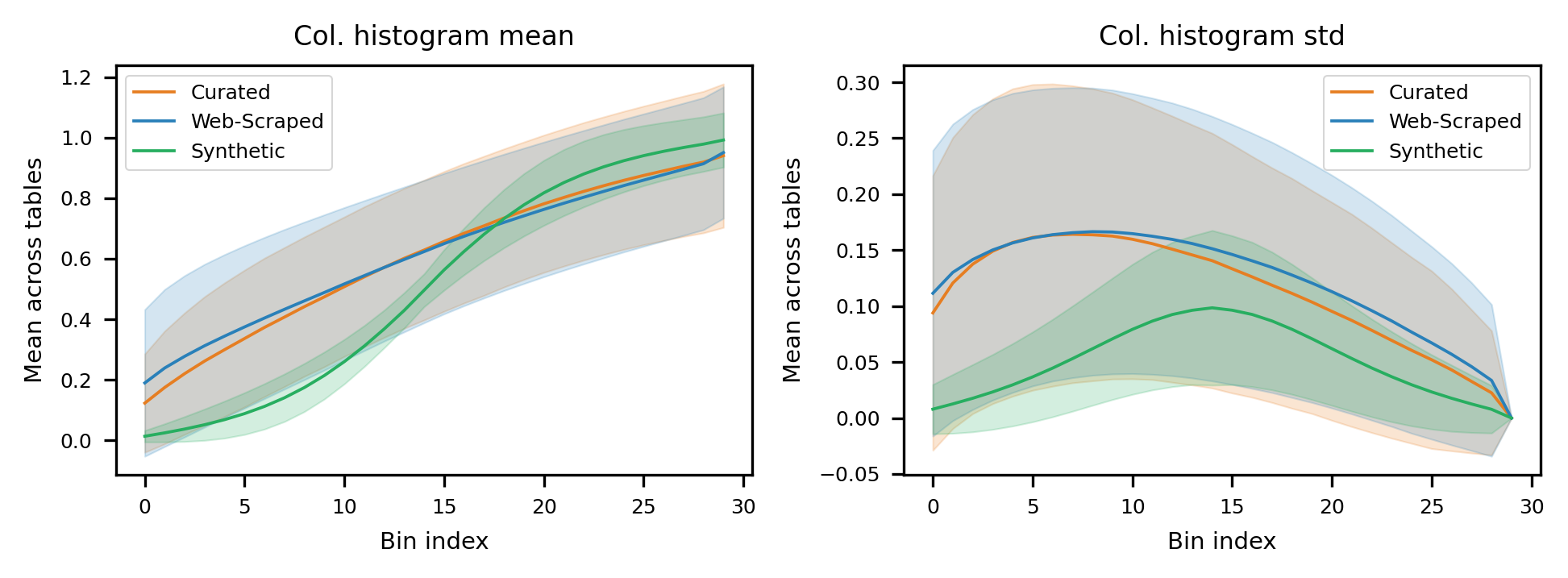}
    \caption{Example visualisations of our aggregate column histogram features. Left: Mean bin values across tables. Right: Mean bin deviations across tables.}
    \label{fig:pairwise-hist-features}
\end{figure}

\clearpage
\newpage

\section{Optimisation}
\label{app:optimisation}

\subsection{Search Space $\theta$}

The search space $\theta$ is defined over the following parameters, partitioned by their conditional relevance to the SCM prior type $\tau$:

\textbf{Shared} (active for all $\tau$):
\begin{itemize}
    \item $\lambda_c \in \{2, \ldots, 20\}$ --- maximum number of target classes.
    \item $p_{\text{cat}} \in [0,\, 0.6]$ --- probability of generating a categorical column.
    \item $p_{\text{ord}} \in [0,\, 1]$ --- probability of ordered multiclass targets.
    \item $\textit{balanced} \in \{0, 1\}$ --- whether class balance is enforced.
    \item $\textit{replay\_small} \in \{0, 1\}$ --- whether small-dataset replay is used.
    \item $\tau \in \{\texttt{mlp\_scm},\, \texttt{tree\_scm},\, \texttt{mix\_scm}\}$ --- SCM prior type.
\end{itemize}

\textbf{Conditional on} $\tau \in \{\texttt{tree\_scm},\, \texttt{mix\_scm}\}$:
\begin{itemize}
    \item $m \in \{\texttt{DT},\, \texttt{ET},\, \texttt{RF},\, \texttt{XGB}\}$ --- tree model family.
    \item $\lambda_d \in [0.3,\, 1.0]$ --- tree depth rate parameter.
    \item $\lambda_n \in [0.3,\, 1.0]$ --- number-of-estimators rate parameter.
\end{itemize}

\textbf{Conditional on} $\tau = \texttt{mix\_scm}$ only:
\begin{itemize}
    \item $p_{\text{mlp}} \in [0,\, 1]$ --- MLP mixing probability (the tree weight is $1 - p_{\text{mlp}}$).
\end{itemize}

Two additional parameters are fixed per dataset rather than searched: the column-count bounds $(\min_f, \max_f)$, derived from the range of $|C|$ observed across the cached real-table features.

\subsubsection{Bayesian, Genetic Searches - AUC}
We begin our attempts at optimisation using bayesian and genetic optimisation approached.

\textbf{Bayesian optimisation} We use the Tree-structured Parzen Estimator (TPE) as implemented in Optuna \cite{akiba_optuna_2019}. At each trial, TPE fits two kernel density estimates over previously evaluated configurations: $\ell(\theta)$ over the top-performing quartile and $g(\theta)$ over the remainder. The next candidate maximises the expected improvement proxy $\ell(\theta)/g(\theta)$. We run $N_{\text{trials}} = 1000$ trials per source dataset.

\textbf{Genetic algorithm} A population of $N_{\text{pop}} = 16$ candidate configurations is evolved over $N_{\text{gen}} = 100$ generations. Each generation proceeds follows the process

\begin{enumerate}
    \item \textbf{Evaluation}: compute $\text{AUC}(\theta_i)$ for all $i$.
    \item \textbf{Elitism}: retain the $N_{\text{elite}} = 4$ fittest individuals unchanged.
    \item \textbf{Selection}: fill remaining slots via tournament selection with $k = 3$ competitors; the individual with the lowest AUC wins.
    \item \textbf{Crossover}: pairs of selected parents produce offspring by sampling each parameter uniformly from either parent.
    \item \textbf{Mutation}: each parameter is independently mutated with probability $p_{\text{mut}} = 0.5$. Continuous parameters receive Gaussian noise $\mathcal{N}(0,\, 0.1 \cdot (\text{hi} - \text{lo}))$; integer parameters use $\mathcal{N}(0,\, (\text{hi} - \text{lo})/6)$; categorical parameters are resampled uniformly.
\end{enumerate}

\begin{figure}[ht]
    \centering
    \includegraphics[width=\linewidth]{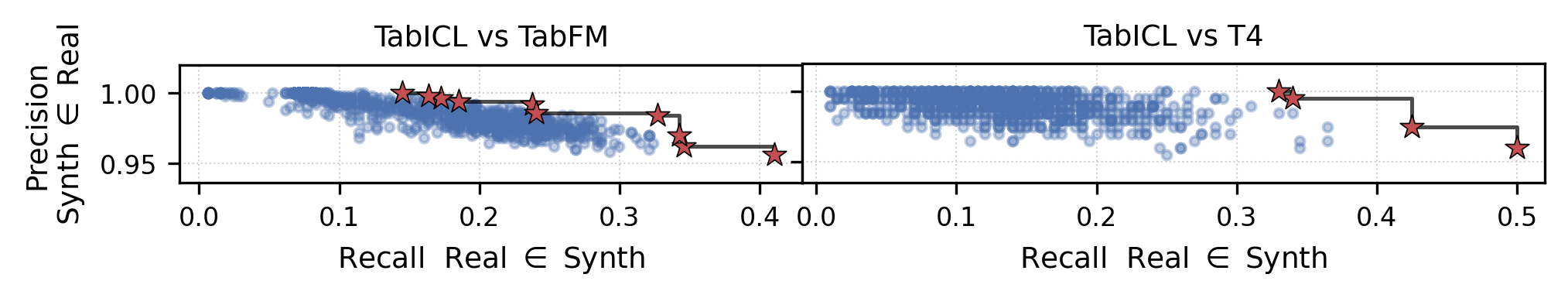}
    \caption{Recall and Precision for optimising the ICL prior to match the FM and T4 datasets, towards the loss function defined in Equation~\ref{eqn:triple-loss}. The Recall/Precision Pareto front is marked with red stars.}
    \label{fig:pareto-fronts}
\end{figure}

\textbf{Results} Neither the Bayesian nor genetic optimisation is able to significantly decrease the AUC of a discriminator, despite extensive experiments and attempts at finding superior optimisation performance.
Though both methods have limited guarantees on convergence, their failure does suggest that the prior cannot produce tables that match those from T4.
To ensure this finding is not a product of our optimisation processes, we conduct a high-resolution grid search over the prior space.


\subsubsection{Bayesian Search - Coverage + AUC}
\label{app:opt-triple}

Given that all optimisations over AUC failed, we also experiment with a reward function that includes coverage alongside AUC through a cartesian combination as in Equation~\ref{eqn:triple-loss}:

\begin{equation}
\label{eqn:triple-loss}
    \mathbf{L} = \left(1/\sqrt{3}\right) \cdot \sqrt{\left(A\in B\right)^2 + \left(B \in A\right)^2 + \left(2 \cdot |0.5 - \textrm{AUC}|\right)^2}
\end{equation}


A grid search is less tractable here, as coverage is expensive to calculate compared to AUC, so we limit our methods to a Bayesian search.
We use 50 Bayesian optimisers, each of 20 trials, as higher trial counts did not lead to improved convergence in our analysis over AUC.
We also use larger evaluation counts of 500 prior-generated tables to reduce aleatoric noise.
A visualisation of Recall and Precision for each trial step can be found in Figure~\ref{fig:pareto-fronts}.

We find that while optimising this three-term reward function does increase recall (Real $\in$ Synth), AUC remains high, and all three metrics degrade when re-run with the same prior parameters and larger sample sizes for the best-performing runs (see Table~\ref{tab:combined-results}).
We note that there is a Pareto front along the Recall/Precision metrics; the best runs have higher recall only with the penality of reduced precision.


\subsection{Further Figures}

\begin{figure}[ht]
    \centering
    \includegraphics[width=\linewidth]{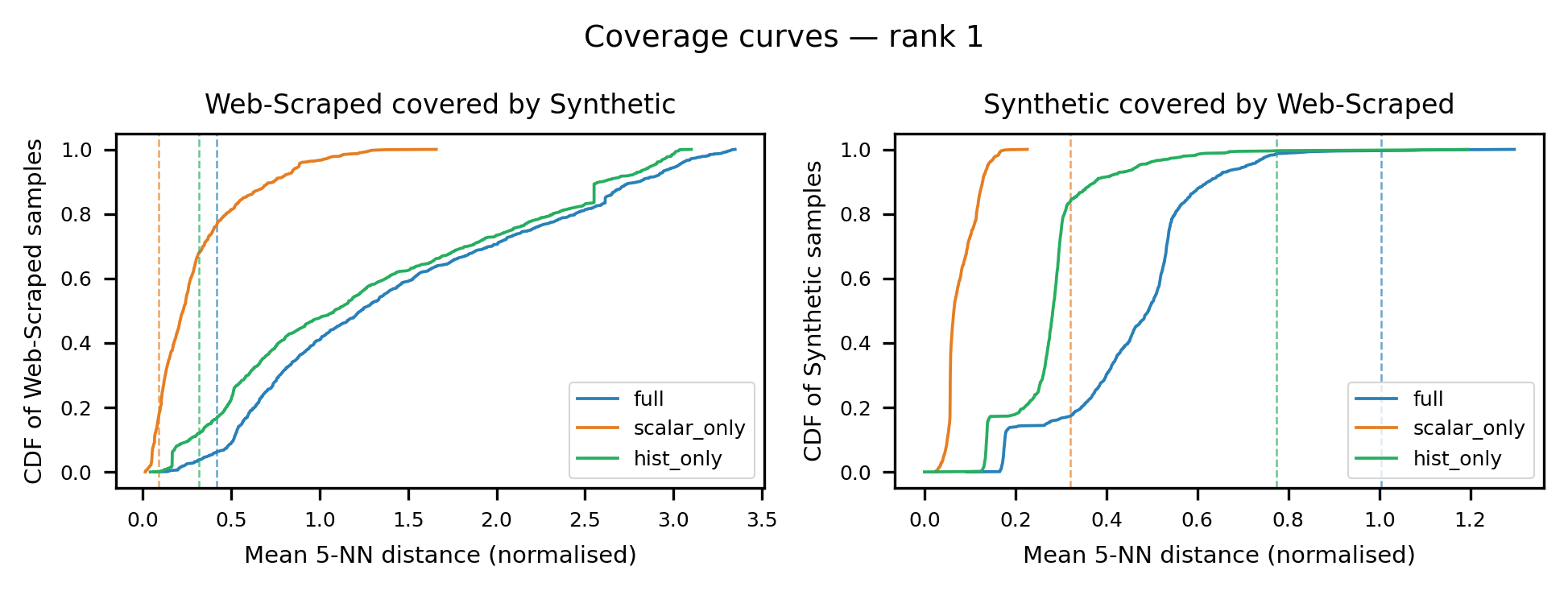}
    \caption{Cumulative coverage curves for ICL over T4 (left) and T4 over ICL (right).}
    \label{fig:grid-coverage}
\end{figure}

\begin{figure}[ht]
    \centering
    \includegraphics[width=\linewidth]{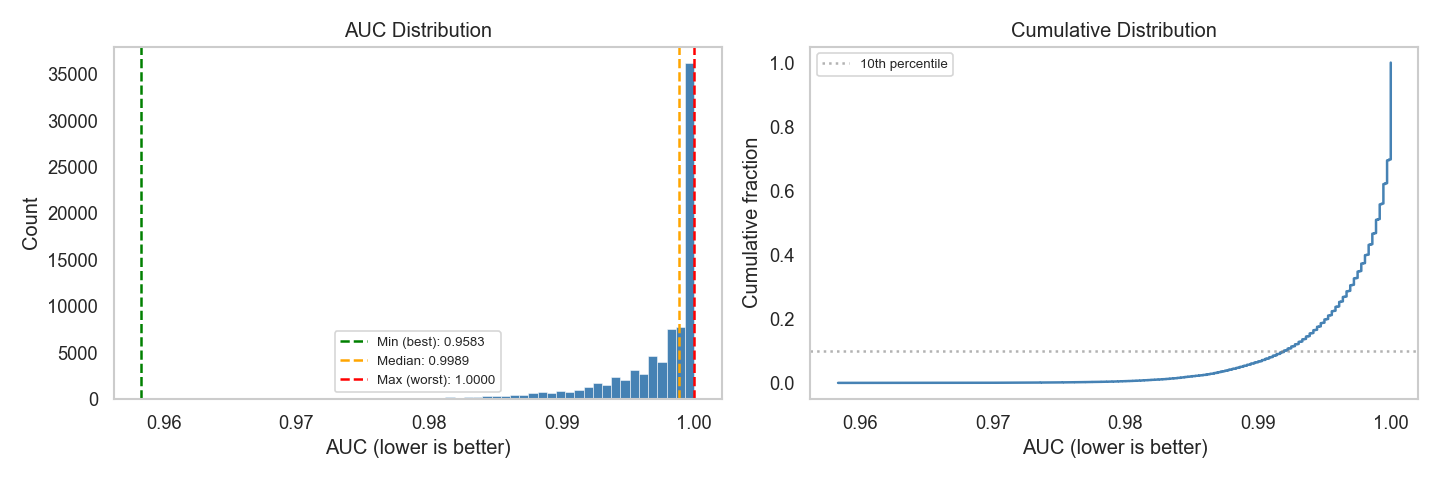}
    \caption{A histogram and a cumulative histogram of AUC values over our grid search.}
    \label{fig:grid-auc}
\end{figure}


\clearpage
\newpage

\subsection{Column Examples}

\begin{figure}[hbtp]
    \centering
    \includegraphics[width=\linewidth]{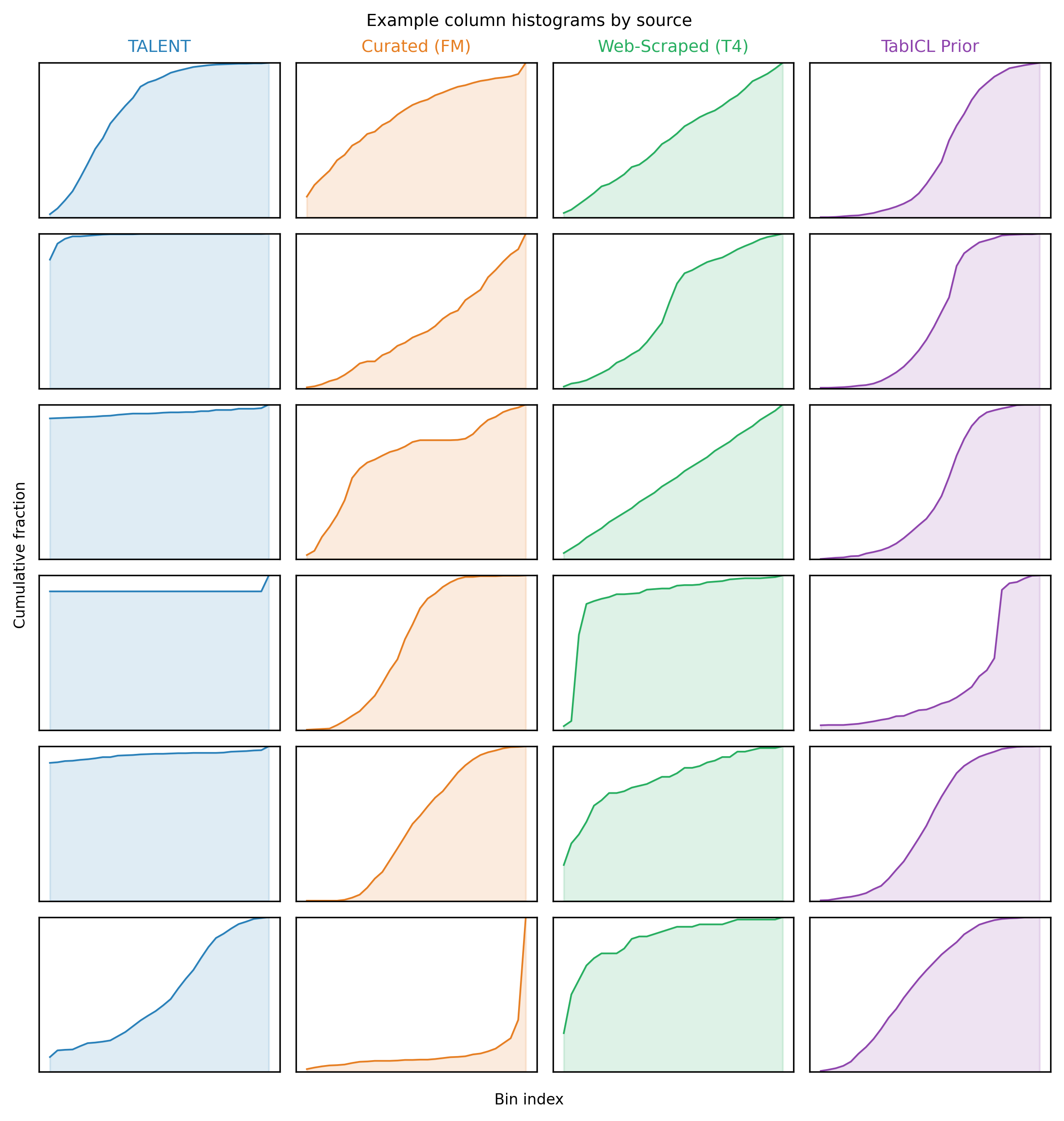}
    \caption{Examples of cumulative histograms over columns randomly selected from the TALENT benchmarks, the curated FM dataset, the web-scraped T4 dataset, and the synthetic prior-based ICL dataset.}
    \label{fig:column-examples}
\end{figure}

\clearpage
\newpage

\section{Benchmarks: Classical Models}
\label{sec:classical-models}

We benchmark TabICL against a suite of classical and automated machine
learning baselines on all binary and multi-class classification datasets in
the TALENT benchmark~\cite{liuTalentTabularAnalytics2025}.  Each model is evaluated under identical
preprocessing and, where applicable, averaged over 3 random seeds.
All models receive the full training data.
For TabICL this means that the maximum amount of training data is presented as context.

\subsection{Gradient-Boosted Trees}

We include three gradient-boosted decision tree methods:
XGBoost~\cite{chenXGBoostScalableTree2016}, LightGBM~\cite{keLightGBMHighlyEfficient2017}, and
CatBoost~\cite{dorogushCatBoostGradientBoosting2018}.  All three build an additive ensemble of
shallow trees by greedily minimising a regularised loss, differing primarily
in their tree-growth strategy and regularisation scheme.  Each ensemble
contains up to 2\,000 trees, with early stopping applied on a held-out
validation split to prevent overfitting.

\subsection{Random Forest}

We include a Random Forest~\cite{breimanRandomForests2001} of 2\,000 trees.
Unlike boosting methods, each tree is trained independently on a bootstrap
sample of the data with a random subset of features considered at each
split, and predictions are aggregated by majority vote.  Trees are capped at
depth 12.

\subsection{Logistic Regression}

Logistic Regression with $\ell_2$ regularisation serves as a linear
baseline.  It models the log-odds of each class as a linear function of the
input features, making it well-suited to linearly separable problems and
providing a simple reference point for non-linear methods.

\subsection{Support Vector Machine}

We include a Support Vector Machine~\cite{cortesSupportvectorNetworks1995} with an RBF
kernel, which implicitly maps inputs to a high-dimensional feature space and
finds the maximum-margin separating hyperplane therein.  SVM is particularly
competitive on small, high-dimensional datasets.

\subsection{$k$-Nearest Neighbours}

$k$-Nearest Neighbours ($k$-NN) classifies each test point by
majority vote among its $k=5$ nearest neighbours in the training set,
weighted by inverse distance.  It is a non-parametric method with no
explicit training phase, making it a useful lower bound on the difficulty
of each dataset.

\subsection{AutoGluon}

AutoGluon~\cite{ericksonAutoGluonTabularRobustAccurate2020} is an automated machine learning system
that constructs a multi-layer stacked ensemble, combining gradient-boosted
trees, neural networks, and other base learners selected and weighted
automatically.  It is included as a strong, well-tuned ensemble baseline
that is representative of the current practical state of the art for
tabular classification.

\section{Benchmarking: Alternative Performance Metrics}

\begin{figure}[ht]
    \centering
    \includegraphics[width=\linewidth]{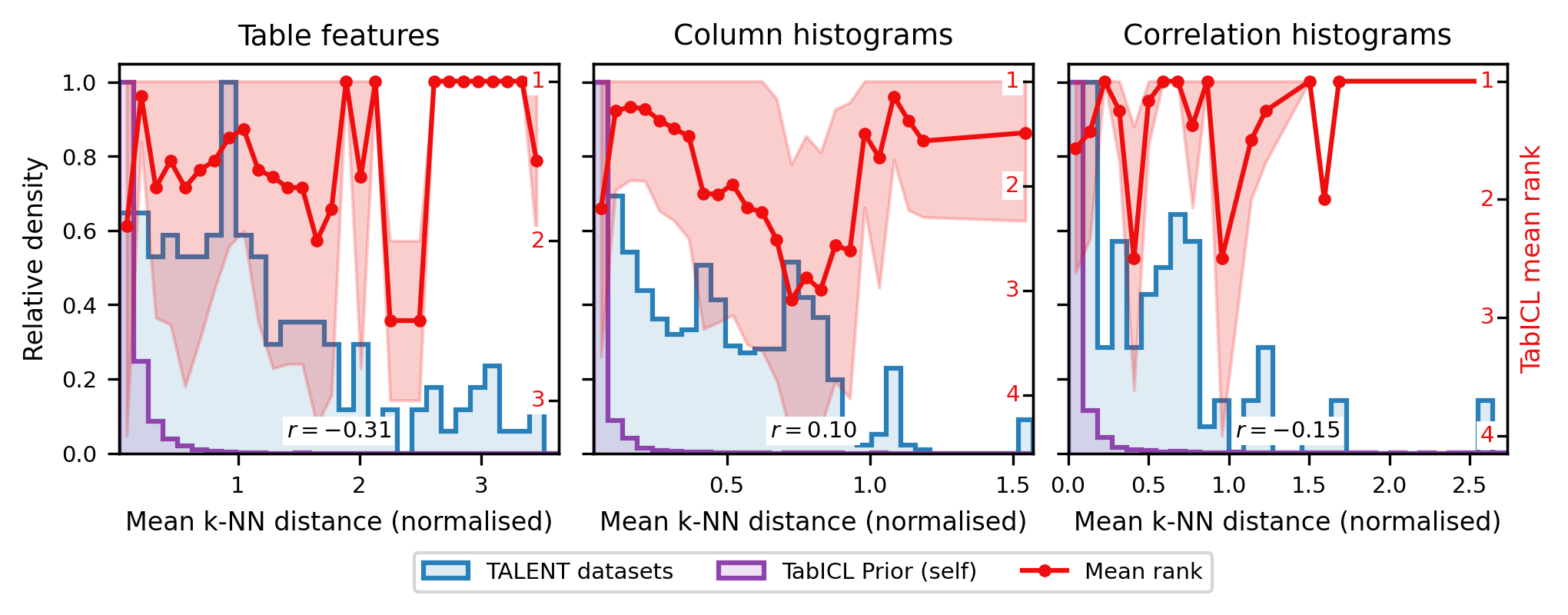}
    \caption{k-NN histograms (5 neighbours) calculated for ICL against itself and the TALENT benchmark tables across our full feature set, cumulative column histograms, and correlation histograms. Plotted alongside is the mean TabICL rank against classical benchmarks for the tables within that bin.}
    \label{fig:rank-vs-coverage}
\end{figure}

\begin{figure}[ht]
    \centering
    \includegraphics[width=\linewidth]{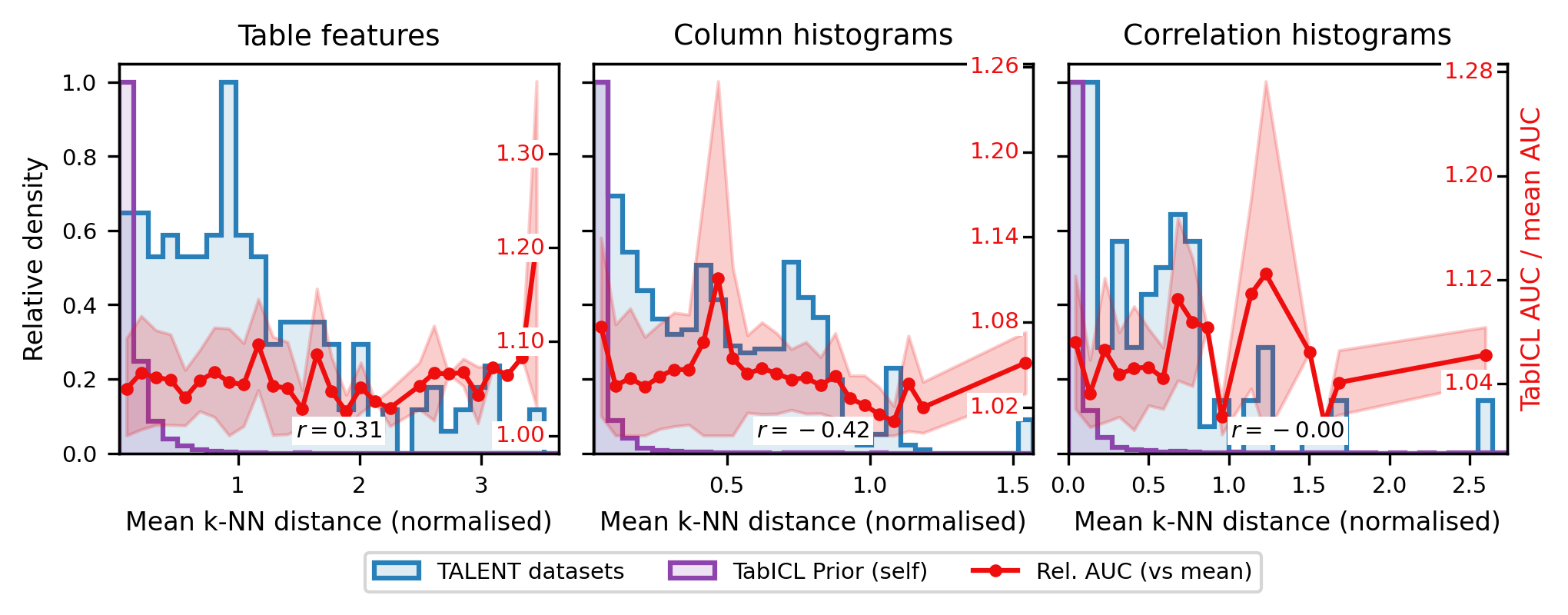}
    \caption{k-NN histograms (5 neighbours) calculated for ICL against itself and the TALENT benchmark tables across our full feature set, cumulative column histograms, and correlation histograms. Plotted alongside is the mean TabICL rank against classical benchmarks for the tables within that bin.}
    \label{fig:mean-auc}
\end{figure}

\begin{figure}[ht]
    \centering
    \includegraphics[width=\linewidth]{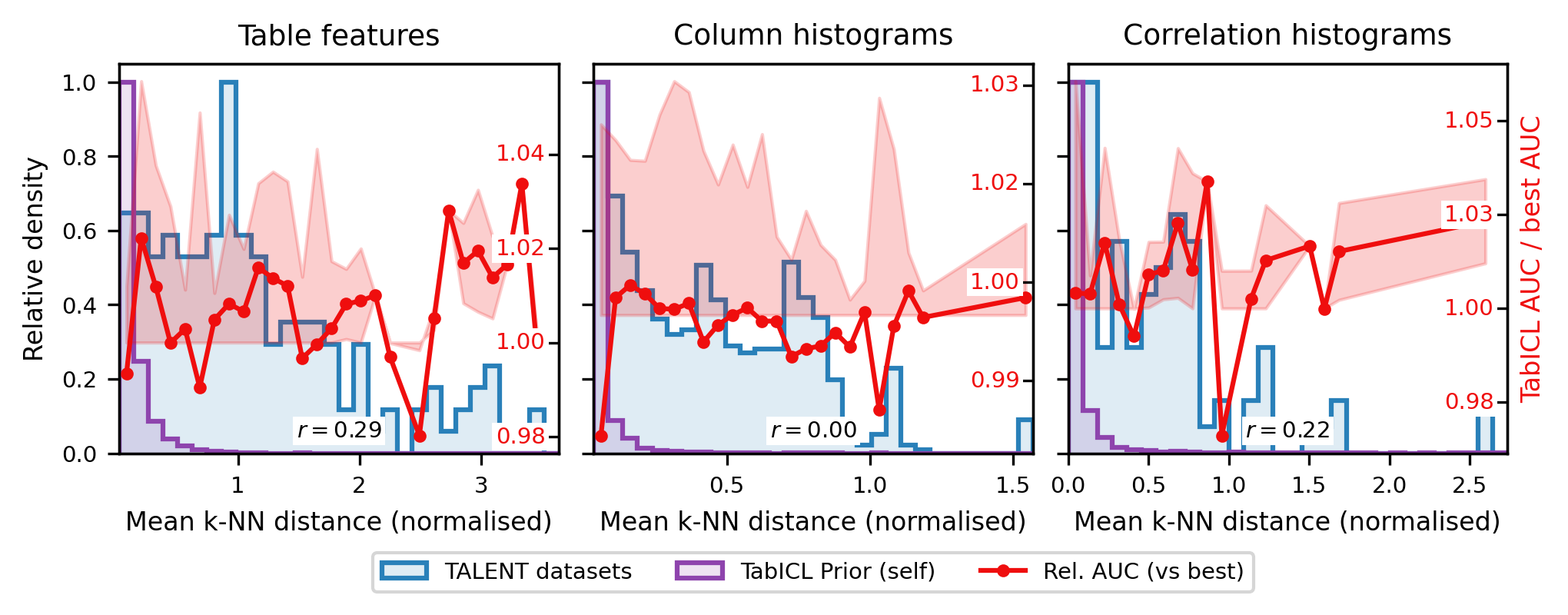}
    \caption{k-NN histograms (5 neighbours) calculated for ICL against itself and the TALENT benchmark tables across our full feature set, cumulative column histograms, and correlation histograms. Plotted alongside is the mean TabICL rank against classical benchmarks for the tables within that bin.}
    \label{fig:max-auc}
\end{figure}

As well as the ranking analysis presented in Figure~\ref{fig:rank-vs-coverage}, we substitute ranking for both the TabICL AUC relative to the mean benchmark performance:
\begin{equation}
    r_{\textrm{AUC}} = \frac{\textrm{AUC}_{\textrm{ICL}}}{\sum_{i\in \textrm{Benchmarks}} \textrm{AUC}_{\textrm{i}}}
\end{equation}
and the TabICL AUC relative to the best performing benchmark:
\begin{equation}
    r'_{\textrm{AUC}} = \frac{\textrm{AUC}_{\textrm{ICL}}}{\textrm{max}\left( \{ \textrm{AUC}_{\textrm{i}} | i \in \textrm{Benchmarks} \} \right)}.
\end{equation}
The mean relative AUC metric is shown in Figure~\ref{fig:mean-auc}, and AUC relative to benchmark maximum in Figure~\ref{fig:max-auc}.
Here we see the same lack of pattern, and low correlations, that we observed when using the rank of the TabICL compared to the benchmark.
These alternative metrics therefore corroborate our findings in the main text.

\subsection{TabArena}
\label{sec:tabarena}

\begin{figure}[ht]
    \centering
    \includegraphics[width=\linewidth]{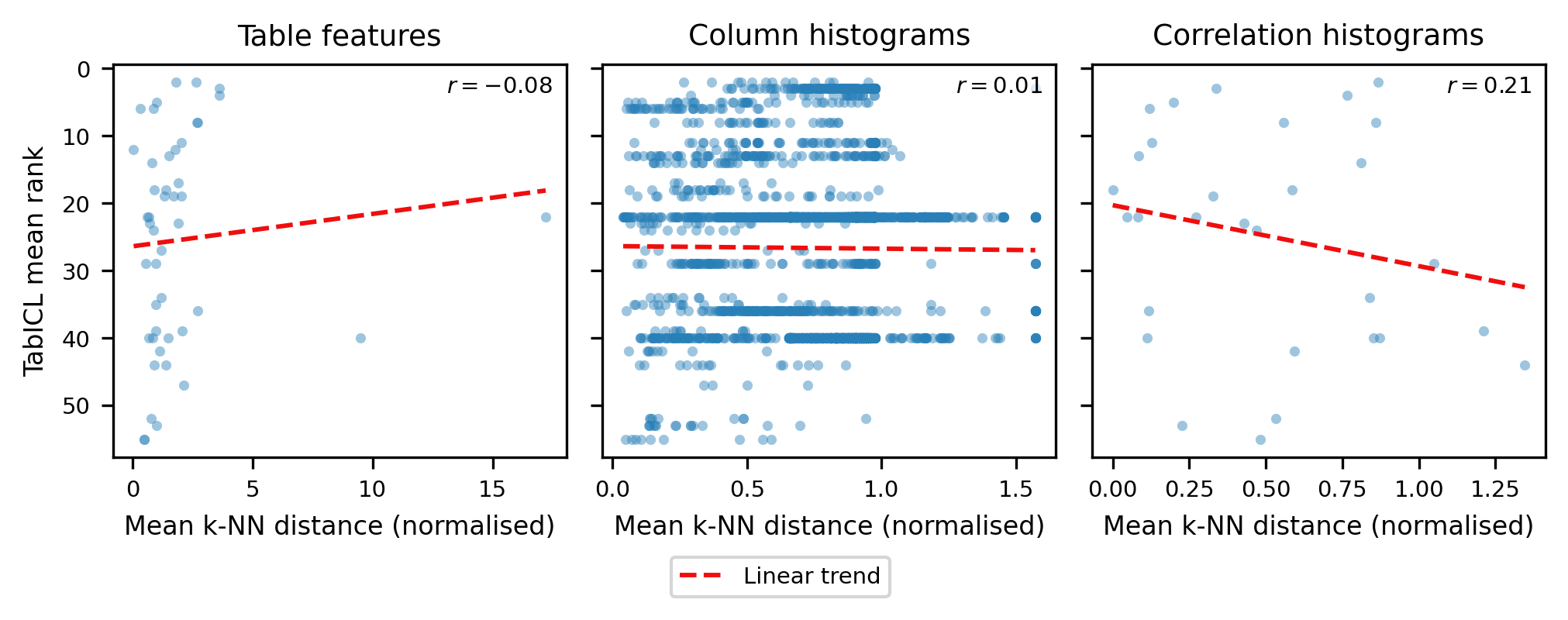}
    \caption{The average rank of TabICL on TabArena, scattered against the average k-NN distance for the benchmark table in question.}
    \label{fig:tabarena-rank}
\end{figure}

\begin{figure}[ht]
    \centering
    \includegraphics[width=\linewidth]{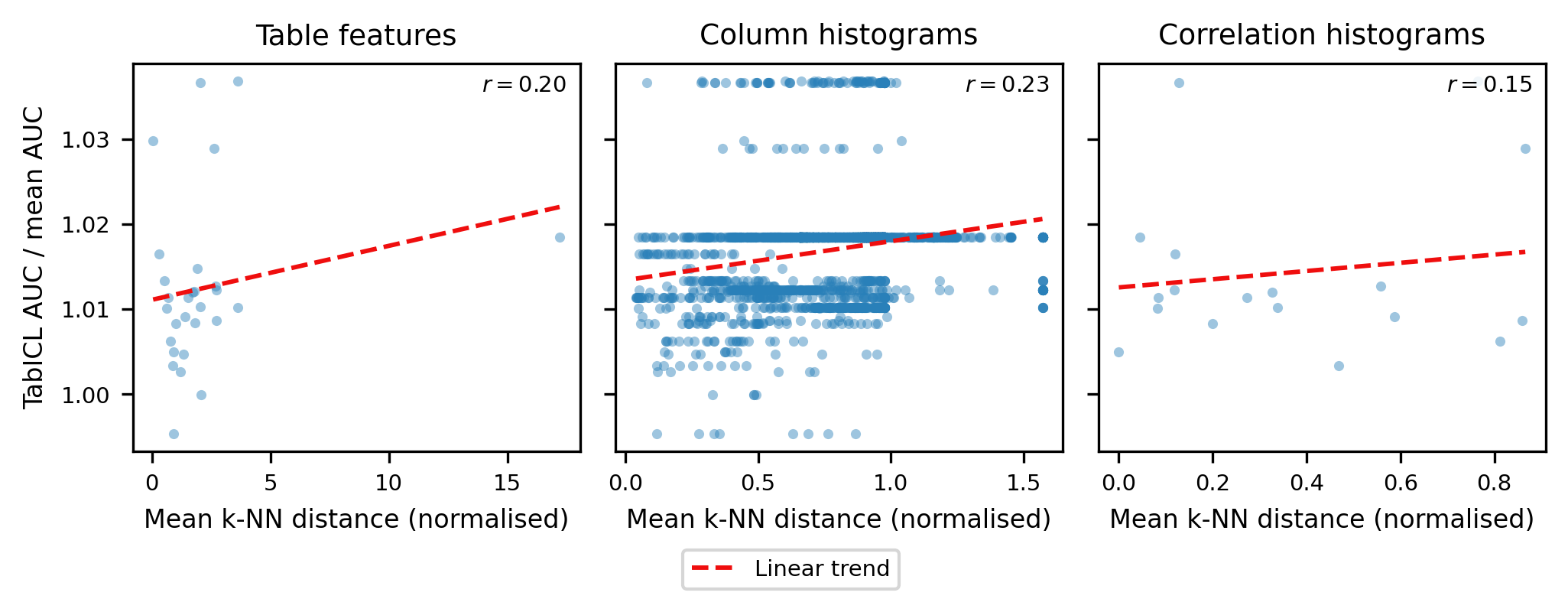}
    \caption{The mean relative AUC of TabICL on TabArena, scattered against the average k-NN distance for the benchmark table in question.}
    \label{fig:tabarena-rel-auc}
\end{figure}

\begin{figure}[ht]
    \centering
    \includegraphics[width=\linewidth]{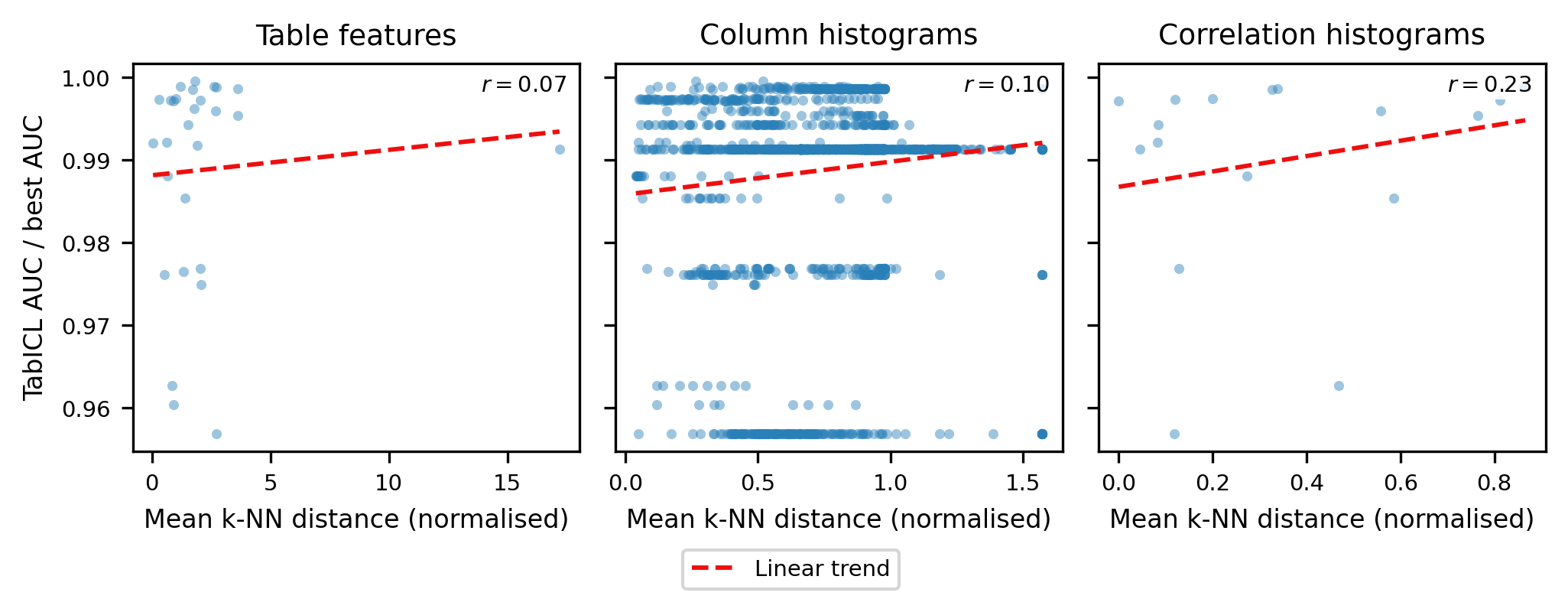}
    \caption{The worst-case relative AUC of TabICL on TabArena, scattered against the average k-NN distance for the benchmark table in question.}
    \label{fig:tabarena-best-auc}
\end{figure}

We perform the same analysis using the pre-compiled ranking results from TabArena, a smaller set of tabular benchmarks.
We visualise these results in Figures~\ref{fig:tabarena-rank},~\ref{fig:tabarena-rel-auc},~\ref{fig:tabarena-best-auc}.
This broadly reproduces the results of our above work; there is little trend between proximity to the prior and performance on these benchmarks.

\subsection{Confound Analysis}
\label{sec:confound}

Dataset-level covariates such as size, dimensionality, and class balance may independently affect both proximity to the synthetic prior and task difficulty, potentially confounding the null result reported in Section~\ref{sec:rq4}.
To address this, we recompute the correlation between k-NN distance and each performance metric using partial correlation, controlling for these covariates.
We consider two covariate sets: a \emph{universal} set  (number of rows, number of columns, and class balance) available for all TALENT classification datasets, and a \emph{full} set that additionally includes skewness, kurtosis, and categorical ratio, available for datasets where these are fully observed.

Results are reported in Table~\ref{tab:partial_corr}.
No partial correlation reaches significance at $p \leq 0.05$ under either covariate set, across all six feature representations and all three performance metrics.
The partial correlations are uniformly small in magnitude, with no consistent sign, and in several cases are smaller in absolute value than the uncorrected correlations in Table~\ref{tab:rank}.
The null result of Section~\ref{sec:rq4} therefore does not appear to be an artefact of covariate confounding.

\begin{table}[ht]
\centering
\caption{Partial correlations between k-NN distance ($k=5$) and performance metrics, controlling for dataset-level covariates. \emph{Universal} covariates (number of rows, columns, and class balance) are available for all TALENT classification datasets; \emph{full} covariates additionally include skewness, kurtosis, and categorical ratio. No result is statistically significant at $p \leq 0.05$.}
\label{tab:partial_corr}
\begin{tabular}{ll rr rr rr}
\toprule
 & & \multicolumn{2}{c}{Rank} & \multicolumn{2}{c}{$\frac{\textrm{AUC}_{\textrm{ICL}}}{\textrm{mean}(\textrm{AUC}_{\textrm{bench}})}$} & \multicolumn{2}{c}{$\frac{\textrm{AUC}_{\textrm{ICL}}}{\textrm{max}(\textrm{AUC}_{\textrm{bench}}})$} \\
\cmidrule(lr){3-4} \cmidrule(lr){5-6} \cmidrule(lr){7-8}
Features & Covariates & $r$ & $p$ & $r$ & $p$ & $r$ & $p$ \\
\midrule
\multirow{2}{*}{Table features}
 & Universal & {$\downarrow \quad$ 0.071} & 0.364 & {$\uparrow \quad$ 0.094} & 0.234 & {$\uparrow \quad$ 0.079} & 0.316 \\
 & Full      & {$\downarrow \quad$ 0.133} & 0.248 & {$\downarrow \quad$$-$0.127} & 0.270 & {$\downarrow \quad$$-$0.046} & 0.691 \\
\multirow{2}{*}{Col. Hists.}
 & Universal & {$\downarrow \quad$ 0.097} & 0.219 & {$\downarrow \quad$$-$0.024} & 0.761 & {$\downarrow \quad$$-$0.007} & 0.928 \\
 & Full      & {$\downarrow \quad$$-$0.039} & 0.734 & {$\downarrow \quad$$-$0.084} & 0.470 & {$\downarrow \quad$$-$0.020} & 0.861 \\
\multirow{2}{*}{Corr. Hists.}
 & Universal & {$\downarrow \quad$ 0.131} & 0.216 & {$\downarrow \quad$$-$0.021} & 0.845 & {$\downarrow \quad$$-$0.021} & 0.841 \\
 & Full      & {$\downarrow \quad$$-$0.161} & 0.286 & {$\downarrow \quad$$-$0.002} & 0.988 & {$\uparrow \quad$ 0.078} & 0.607 \\
\midrule
\multirow{2}{*}{TabICL Columns}
 & Universal & {$\downarrow \quad$ 0.117} & 0.137 & {$\uparrow \quad$ 0.077} & 0.325 & {$\uparrow \quad$ 0.042} & 0.596 \\
 & Full      & {$\downarrow \quad$ 0.021} & 0.856 & {$\uparrow \quad$ 0.057} & 0.620 & {$\downarrow \quad$$-$0.031} & 0.789 \\
\multirow{2}{*}{TabICL Rows}
 & Universal & {$\downarrow \quad$ 0.141} & 0.072 & {$\downarrow \quad$$-$0.004} & 0.958 & {$\downarrow \quad$$-$0.044} & 0.573 \\
 & Full      & {$\downarrow \quad$ 0.025} & 0.832 & {$\downarrow \quad$$-$0.027} & 0.818 & {$\downarrow \quad$$-$0.027} & 0.814 \\
\multirow{2}{*}{TabICL Concat.}
 & Universal & {$\downarrow \quad$ 0.136} & 0.084 & {$\uparrow \quad$ 0.018} & 0.823 & {$\downarrow \quad$$-$0.029} & 0.709 \\
 & Full      & {$\downarrow \quad$ 0.016} & 0.892 & {$\downarrow \quad$$-$0.012} & 0.915 & {$\downarrow \quad$$-$0.028} & 0.807 \\
\bottomrule
\end{tabular}
\end{table}

\newpage
\clearpage

\section{Future Work Brief}
\label{sec:future-work}

The null result of RQ4 admits two candidate explanations: that in-context learning is inherently
robust to pre-training distributional shift by virtue of inference-time conditioning on task-specific
examples, or that the inductive biases instilled by SCM-based pre-training transfer broadly
regardless of low-level distributional alignment. Directly testing these accounts requires
a controlled pre-training ablation that is computationally prohibitive at the scale of the
present work; we outline the design here for completeness.

\subsection*{Experimental Design}

The ablation requires training two model variants on deliberately degraded priors.
The first, a \textit{distribution-matched prior}, preserves SCM causal structure but replaces
synthetic marginal column distributions with those drawn from real tables, for instance by
fitting empirical CDFs over T4 or FM columns and using these as leaf-node noise distributions.
The second, a \textit{structure-ablated prior}, retains realistic marginal distributions but
removes causal structure by permuting DAG edges, breaking generative dependencies between
features while preserving column-level realism. Both variants should otherwise follow the
TabICL training setup exactly.

Each variant is evaluated on the TALENT benchmark under the proximity-performance
correlation analysis of Section~\ref{sec:rq4}, with internal embeddings from each trained
model used as the proximity measure, following Table~\ref{tab:rank}.

\subsection*{Predicted Outcomes}

If the null result is driven by the causal inductive biases of SCM pre-training, it should
survive in the distribution-matched variant but degrade in the structure-ablated one.
If robustness instead arises from the inference-time conditioning mechanism of ICL,
the null result should hold under both. A third outcome, degradation under both
ablations, would suggest that causal structure and distributional realism are jointly
necessary, motivating a more fine-grained factorial design.

\subsection*{Computational Scope}

Each variant requires a full pre-training run comparable in cost to the original TabICL
training, and a minimum of two variants plus a matched control are required for a clean
comparison. This places the experiment beyond the scope of the present work. We leave
direct investigation to future work, and note that the mechanistic question it
addresses, whether distributional coverage or causal inductive bias is the primary
driver of TFM generalisation, has direct implications for prior design more broadly.


\end{document}